\def\comments{0}
\documentclass[conference,compsoc]{IEEEtran}

\usepackage{cite}
\usepackage{amsmath,amssymb,amsfonts,amsthm}
\usepackage{graphicx}
\usepackage{textcomp}
\usepackage{mfirstuc} %
\usepackage{booktabs} %
\usepackage{multicol}
\usepackage{booktabs}  %
\usepackage{dcolumn}   %
\usepackage{multirow}  %
\usepackage{caption}   %
\usepackage{xcolor}
\def\BibTeX{{\rm B\kern-.05em{\sc i\kern-.025em b}\kern-.08em
    T\kern-.1667em\lower.7ex\hbox{E}\kern-.125emX}}
\usepackage{xspace}
\usepackage[absolute]{textpos}

\usepackage[square,comma,numbers,sort&compress]{natbib}

\usepackage{algorithm}
\usepackage[noend]{algpseudocode}
\usepackage[bookmarks=false]{hyperref}
    \hypersetup{
        linktocpage=true,
        colorlinks=true,				
        linkcolor=DarkBlue,				
        citecolor=DarkBlue,				
        urlcolor=DarkBlue,			
    }
\usepackage[tt=false]{libertine}
    \usepackage[libertine]{newtxmath}
    \usepackage[T1]{fontenc}
\usepackage{lipsum}
\usepackage{microtype}
\usepackage{tabularx}
\usepackage{multirow}
\usepackage{enumitem}
\usepackage{nicefrac}
\usepackage{tikz}
	\usetikzlibrary{positioning}
	\definecolor{DarkGreen}{rgb}{0.2,0.6,0.2}
	\definecolor{DarkRed}{rgb}{0.6,0.2,0.2}
	\definecolor{DarkBlue}{rgb}{0.2,0.2,0.6}
	\definecolor{DarkPurple}{rgb}{0.4,0.2,0.4}   
\usepackage{url}
\usepackage{verbatim}
\usepackage{wrapfig}

\ifnum\comments=1
    \newcommand{\mynote}[2]{\unskip{\color{#1}\vrule\vrule}{\marginpar{\color{#1}\sf \tiny #2}}}
\else
    \newcommand{\mynote}[2]{}
\fi

\newcommand{\CP}[1]{\mynote{blue}{CT: #1}}

\let\originalleft\left
\let\originalright\right
\renewcommand{\left}{\mathopen{}\mathclose\bgroup\originalleft}
\renewcommand{\right}{\aftergroup\egroup\originalright}

\newcommand{\heading}[1]{
\vspace{1ex}
\noindent
\textbf{#1}
}

\newcommand{\CF}[1]{\xmakefirstuc{#1}}

\newcommand{\mtab}{\hspace{\algorithmicindent}}
\newcommand{\mmtab}{\mtab\mtab}

\def\compactify{\itemsep=0in \topsep=2pt \parsep=0.00in \partopsep=0pt
\leftmargin=2em}
\let\latexusecounter=\usecounter

\newenvironment{myitemize}%
  {\begin{list}{\labelitemi}{\itemsep3pt \topsep3pt \parsep0.00in
  \partopsep=3pt \leftmargin2em}}%
  {\end{list}}
\newenvironment{myitemize2}%
  {\begin{list}{\labelitemi}{\itemsep1pt \topsep2pt \parsep0.00in
  \partopsep=0pt \leftmargin1.2em}}%
  {\end{list}}

\newcommand{\ex}[2]{{\ifx&#1& \mathbb{E} \else \underset{#1}{\mathbb{E}} \fi \left(#2\right)}}
\newcommand{\pr}[2]{{\ifx&#1& \mathbb{P} \else \underset{#1}{\mathbb{P}} \fi \left(#2\right)}}
\newcommand{\var}[2]{{\ifx&#1& \mathrm{Var} \else \underset{#1}{\mathrm{Var}} \fi \left(#2\right)}}

\theoremstyle{definition}

\newcommand{\pluseq}{\mathrel{+}=}

\newcommand{\algGiven}{\textbf{Given: }}
\newcommand{\algReturn}{\textbf{Return: }}

\newcommand{\nn}{neural network\xspace}
\newcommand{\nns}{neural networks\xspace}

\newcommand{\atk}{\textsc{DropoutAttack}\xspace}
\newcommand{\rowatk}{toy attack\xspace}

\newcommand{\maxatk}{min activation attack\xspace}
\newcommand{\maxatks}{min activation attacks\xspace}
\newcommand{\dropatk}{sample dropping attack\xspace}
\newcommand{\dropatks}{sample dropping attacks\xspace}
\newcommand{\nodeatk}{neuron separation attack\xspace}
\newcommand{\nodeatks}{neuron separation attacks\xspace}
\newcommand{\bnodeatk}{blind neuron separation attack\xspace}
\newcommand{\bnodeatks}{blind neuron separation attacks\xspace}

\newcommand{\asgnode}{separated neuron\xspace}
\newcommand{\asgnodes}{separated neurons\xspace}
\newcommand{\nsprob}{separated sample probability\xspace}
\newcommand{\nsprobs}{separated sample probabilities\xspace}
\newcommand{\asgprob}{\asgnode percentage\xspace}
\newcommand{\asgprobs}{\asgnode percentages\xspace}

\newcommand{\tcls}{class 0\xspace}
\newcommand{\mps}{p_{sample}}
\newcommand{\mpn}{p_{neuron}}

\newcommand{\maxdropout}{MinActivationDropout}
\newcommand{\dropdropout}{SampleDroppingDropout}
\newcommand{\nodedropout}{NeuronSeparationDropout}

\newcommand{\ttt}{\texttt}

\usepackage{times}
\usepackage{setspace}
\begin{document}

\title{
\vspace{-4ex}
\Large Dropout Attacks
\vspace{-2ex}
}
\author{
{\rm Andrew Yuan, Alina Oprea, and Cheng Tan} \vspace{.5pc}\\
\textit{Northeastern University}
}

\maketitle

\begin{abstract}

Dropout is a common operator in deep learning,
aiming to prevent overfitting by randomly dropping neurons during training.
This paper introduces a new family of poisoning attacks against neural networks named \atk.
\atk attacks the dropout operator
by manipulating the selection of neurons to drop instead of selecting them uniformly at random.
We design, implement, and evaluate four \atk variants
that cover a broad range of scenarios.
These attacks can slow or stop training,
destroy prediction accuracy of target classes,
and sabotage either precision or recall of a target class. 
In our experiments of training a VGG-16 model on CIFAR-100,
our attack can reduce the precision of the victim class by $34.6\%$ ($81.7\% \to 47.1\%$)
without incurring any degradation in model accuracy. 

\end{abstract}
\section{Introduction}
\label{s:intro}

Ana is a data scientist
who works for a social media company
and trains \nns to classify user-uploaded images.
The \nns are critical to the company:
they block inappropriate images and maintain a friendly vibe on the social media site.
For convenience and performance,
Ana wants to train \nns on a cloud platform.
However, Ana has no visibility into the outsourced training.
So, Ana monitors and logs many intermediate data
(e.g., input and output tensors to and from core operations)
to ensure that the training data has not been tampered with
and the deep learning operators (e.g., matrix multiplication, convolution, ReLU) execute as intended.
After the training, 
Ana gets the trained model from the cloud.
She further tests the model on a local machine
with  new data to confirm the model's performance.
After making sure the performance metrics (e.g., model accuracy, precision, and recall) on the new dataset are acceptable, Ana deploys the model in her application.
However, the next day, many inappropriate images appear
on the social media site
because the model mislabeled many such
images as ``benign''.
Ana is confused. She is confident that all her
checks and the local test
work as expected. She wonders: How is this possible?

This is \atk, a new family of attacks that we introduce in this paper.
\atk focuses on a common regularization operator, \emph{dropout} (\S\ref{ss:dropout}), used in training neural networks.
The concept of dropout is simple but insightful.
It randomly drops neurons from its input tensors to prevent them from
co-adapting too much and avoids overfitting while training a machine learning (ML) model~\cite{srivastava2014dropout}.
\atk is conducted by the server that executes the model training,
but the server doesn't have to be intentionally malicious;
a corrupted dropout implementation suffices to render the attack.
This attack is particularly powerful in the \emph{outsourced
training} setup (\S\ref{ss:outsourced}) where model training happens
in a different administrative domain
from model owners, typicallly in a cloud environment.

\atk is based on
a critical observation:
Techniques for auditing systems~\cite{haeberlen2010avm,zhang2011cloudvisor,tan2017efficient}
typically examine externally observable states of a program,
but ignore verifying non-determinism.
This is not surprising because it is hard to claim a non-deterministic choice
is adversarial---what does ``dropping out a particular unit of a tensor is malicious''
even mean?---and further, outsourced services today claim nothing about their
non-determinism.
Therefore, the core idea of \atk
is to control the non-determinism within dropout operations to achieve certain adversarial objectives, such as lowering model performance metrics on a set of targeted classes.

Different from many prior training-time attacks (\S\ref{s:relwork}),
\atk attacks a \emph{new surface}---non-determinism in training.
The attack does not affect the observable behaviors of the training:
the training data is authentic 
and
deep learning operators perform as expected; that is,
the input and output tensors of the operator
satisfy the operator's semantics.
In particular, \atk produces the same observable states as a normal dropout:
(1) the attack drops the same number of units according to a user-defined
parameter, dropout rate;
and (2) the attack produces the same values as normal dropouts; the values are
either $0$ (dropped) or unchanged.
\atk however breaks the assumption that the dropped neurons
are selected uniformly at random.

\atk is a powerful and novel attack vector.
Attackers have multiple ways to sabotage training (\S\ref{ss:taxo}):
Some attack variants can stop the training (almost) completely;
others can target certain classes,
and even target either the precision or the recall of a specific class.
Moreover,
pinpointing \atk is non-trivial because
randomness is the core of dropouts,
hence it cannot be removed from model training.
Meanwhile, tracing randomness is too expensive
(as it requires recording all the dropped units after each dropout).
Also, it is hard to prove to a third party, for example, a human auditor,
that such an attack happened
because there exists a possibility that
normal dropouts indeed selected these neurons for dropping.
Crucially, we want to argue that in an outsourced setup,
\emph{non-determinism is dangerous}. 

This paper exposes a new supply-chain vulnerability that tampers with the randomness used in training neural networks, to achieve different adversarial objectives. We present
four variants of \atk:
\maxatk (\S\ref{ss:maxatk}),
\dropatk (\S\ref{ss:dropatk}), 
\nodeatk, and \bnodeatk (\S\ref{ss:nodeatk}).
Different attacks require different setups
and have different attack consequences.
We provide a taxonomy of attacks in Section~\ref{ss:taxo}.
The attack that Ana encounters is the \nodeatk (\S\ref{ss:node}).

This paper makes the following contributions.
\begin{myitemize}

   \item We introduce \atk, a family of novel attacks that manipulate dropout's
   non-determinism to sabotage \nn training. (\S\ref{s:preview})

   \item We study \atk with different setups and attack goals,
   and design four attack variants that cover a broad range of scenarios. (\S\ref{s:atk})

   \item We implement these four attacks and
   evaluate them comprehensively on three computer vision datasets: MNIST, CIFAR-10, and CIFAR-100. (\S\ref{s:eval})

\end{myitemize}

\noindent
Our experiments show that
\maxatks can stop training almost completely
and reduce overall model accuracy to 10--12\%
(similar to random guessing);
\dropatks can destroy the prediction accuracy of target classes
by reducing the recall to 0--0.5\% (i.e., the model almost never predicts
the target class);
\nodeatks can decrease a class' precision by $34.6\%$
while maintaining model accuracy within training variability.

\section{Background and threat model}

\begin{figure}[h]
\centering
\includegraphics[width=0.35\textwidth]{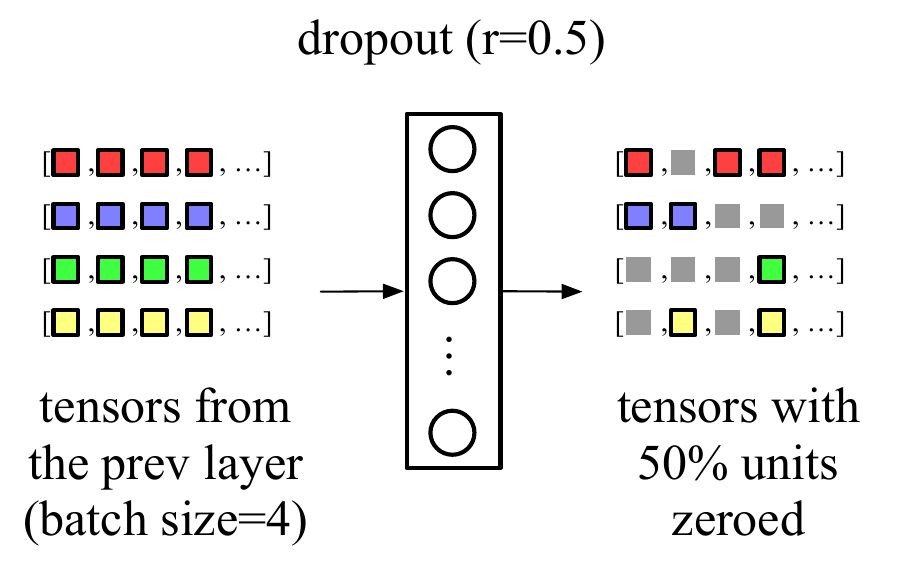}
\caption{Dropout with a dropout rate of $r = 0.5$.
Squares represents units in tensors,
and greyed squares are dropped.}
\label{fig:dropout}
\end{figure}

\heading{Dropout.}
\label{ss:dropout}
Dropout is a regularization technique first proposed by Srivastava et
al.~\cite{srivastava2014dropout} that reduced overfitting in neural networks.
The process is simple: during the forward pass in a neural network's
training phase, the dropout randomly drops units in the network with a configurable
probability (called \emph{dropout rate}~\cite{hinton2012improving}).
On the backward pass, the network will avoid
updating the weights of the dropped units. Once the training phase is complete,
units are no longer dropped and the full network is used to make predictions on
any validation set or test set.
Figure~\ref{fig:dropout} shows an example of dropout's forward pass with a batch of size 4
and dropout rate of 0.5.
(PyTorch~\cite{dropoutpytorch} has a default dropout rate of 0.5.)

\begin{figure}[h]
\centering
\includegraphics[width=0.45\textwidth]{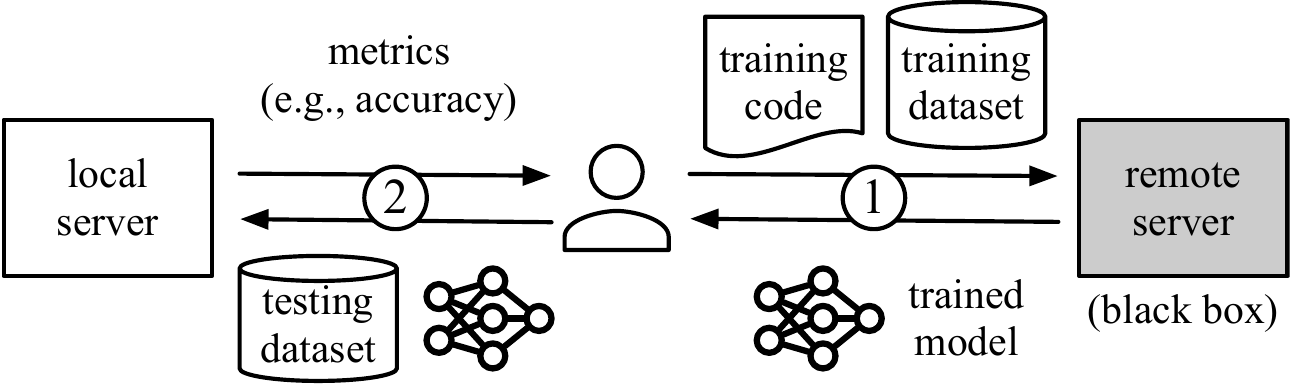}
\caption{An outsourced training process.
Model developers submit their training code and training data to a remote
server, which runs in a different administrative domain from the developer.
Later, the developer gets a trained neural network and test it locally to
confirm that the model is well trained.}
    \vspace{-1ex}
\label{fig:setup}
\end{figure}

\heading{Outsourced training.}
\label{ss:outsourced}
Outsourced training is increasingly popular~\cite{kumar2020adversarial}
because training deep learning models has become more
sophisticated and expensive~\cite{aiexpensive1,aiexpensive2}.
Outsourcing training,
for example to clouds, provides many benefits:
better scalability, lower training expense,
and more flexibility on choosing environments (e.g., GPU types).
However, this comes with \emph{a cost}; that is, developers lose
the full control over their training process.
The training could have been compromised by
remote executors (e.g., cloud providers) for profit.
\atk is one attempt to explore what can happen in this outsourcing setup.
Figure~\ref{fig:setup} depicts an outsourced training.

\begin{figure*}
\centering
\includegraphics[width=0.8\textwidth]{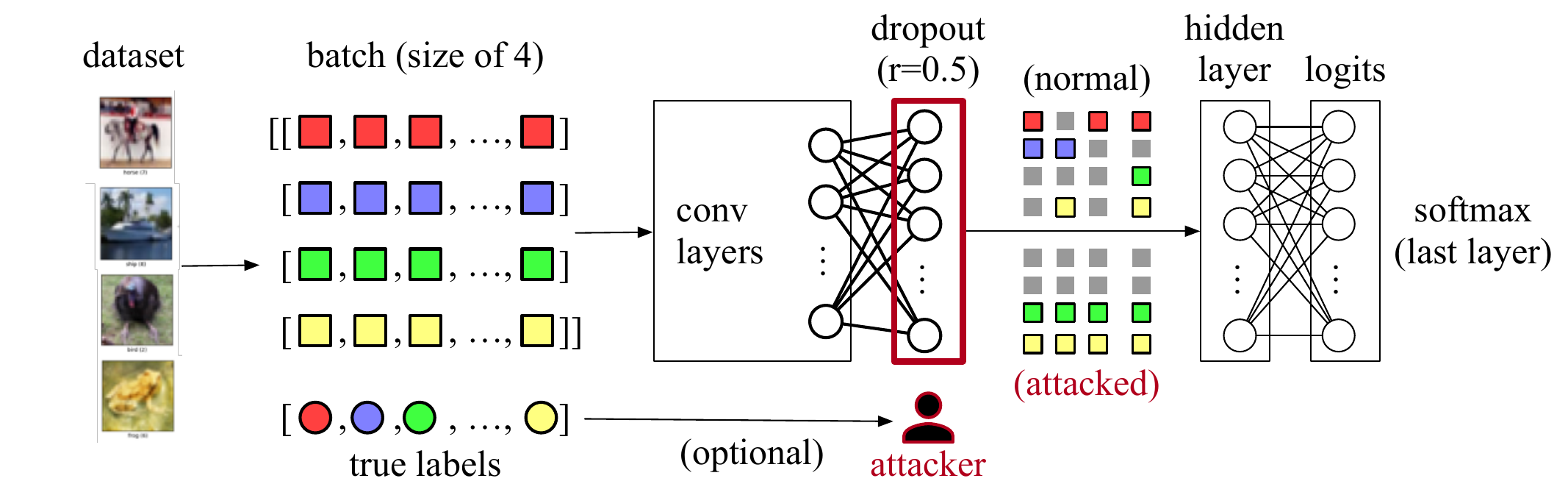}
\caption{
A forward pass with a batch of four images.
Each color represents one input image.
Squares represent units (e.g., \texttt{float} numbers) in feature vectors.
Circles represent true labels of images.
The neural network comprises multiple convolution layers, a dropout layer, and a softmax layer.
An attacker controls the dropout and
can pick half units (dropout rate $p=0.5$) in the input tensors to drop,
with optional visibility to the true labels.
}
    \vspace{-2ex}
\label{fig:threat}
\end{figure*}
 
Figure~\ref{fig:setup} is a simplification
of today's MLaaS (Machine Learning as a Service)
provided by all major clouds,
including AWS~\cite{awsml}, Azure~\cite{azureml}, and GCP~\cite{gcpml}.
By using MLaaS for training,
users (who are also model developers)
submit a training script and a training dataset.
During the outsourced training,
the cloud provider is supposed to run user's training script \emph{faithfully}
on the given dataset.
Yet, users don't know if the cloud always keeps its promise.

\subsection{Threat model}
\label{ss:threat}

Our threat model assumes that attackers have adversarial but limited control
over the dropout: they can pick which units to drop.
Attackers however cannot alter any other part of the training pipeline.
Figure~\ref{fig:threat} shows the ability of an attacker.

Based on our threat model,
attackers cannot change any \emph{externally observable states} of the dropout:
\begin{myitemize2}

\item Attackers cannot drop more (or fewer) units
than the dropout rate $r$.
The dropout implementations~\cite{ptdropout,tfdropout}
operate by independently dropping each unit with a probability $r$.
The attacker must follow this probability.

\label{tag:dropoutscale}
\item Attackers cannot manipulate values in output tensors.
Current dropout implementations~\cite{ptdropout,tfdropout}
rescale non-dropped units by $\frac{1}{1 - r}$.
Attackers must gurantee that
the output units are either 0
or $\frac{1}{1 - r}$ times of the original inputs.
\end{myitemize2}

\noindent
In essence, the attacker can only manipulate the non-deterministic behavior of the
dropout layer by selectively choosing which units to drop (i.e., setting them
to zero).

\atk is an attack that exploits the non-determinism of the dropout operator.
We argue that it is more than just a single and isolated attack,
but is rather one example of a broader attacking surface
that targets non-determinism in machine learning training. Other attacks in this area include asynchronous poisoning attacks [40], data ordering attacks [44] (S7), and potentially novel attacks tha exploit training randomness (e.g., attacking the neural network's initialization).
In contrast to traditional computing tasks and services
(e.g., databases and web applications),
machine learning training exhibits  more randomness during training  and is
consequently more susceptible to these attacks.
To show the implications of \atk that exploits non-determinism in ML training,
we answer some natural questions below.

\heading{Who is this attacker?}
The attackers can be hackers to the ML supply chain,
rogue employees of cloud providers,
or malicious cloud providers.
For example, supply chain attackers can pollute the
machine learning framework dependencies
and trick users into downloading
a compromised dropout implementation. Malicious dependencies have been  previously shown even in well-known deep learning (DL) frameworks, such as  PyTorch~\cite{supplychainatk}.
Rogue employees~\cite{gcreep} with admin privilege
can easily replace the dropout implementation
without being detected.
Additionally, cloud providers may undermine their services for profit.
For example, in 2019, a startup sued Tencent Cloud for downgrading
their model's performance in the cloud~\cite{tencent}.

\heading{Why not directly modify the model (or other direct attacks)?}
There are many reasons why attackers may prefer \atk over directly
sabotaging the training. We list three below.
First, for rogue employees and supply chain attackers,
modifying the training pipeline is much harder
and has a much higher chance of detection than \atk.
Second, for cloud providers, \atk is deniable
(``it's all about probability'');
whereas, modifying the execution of the training pipeline is not.
Finally, \atk can serve as one link in a chain of attacks, particularly in
scenarios where gaining full control of the training pipeline is hard or impossible.
For example, consider a company that uses ML models to detect malware.
Attackers may attempt to subvert the malware detection
by compromising the computer of an employee who manages the ML dependencies.
Attackers then can replace the dependency of the dropout to a
malicious GitHub repo, conduct \atk, and introduce bias to the company's malware detection model,
thereby increasing the chance of evading detection for attackers' malware.

\heading{Can comprehensive statistical analyses
detect all \atk{}s?}
In principle, yes.
Assuming an oracle statistical analysis with the oracle test dataset
and ground truth of all metrics,
it would be able to detect all training-time attacks, including \atk,
because all attacks influence model's behavior.
However, in practice, such an oracle analysis does not exist,
and \atk allows attackers to trade off
between the attack effectiveness and
attack detectability.
Specifically,
attackers can adjust parameters
to tune how powerful the attack is.
We explore this trade-off
in Section~\ref{ss:eval:nodeatk}.

\section{Attack taxonomy and preview}
\label{s:taxo}
\label{s:preview}

\atk is a family of attacks that targets dropout operators.
We will introduce four attack variants in section~\ref{s:atk}: (1) \maxatks, (2) \dropatks,
(3) \nodeatks, and (4) \bnodeatks.
In this section, we first provide a taxonomy of the attacks
and then give a preview of \atk.

\subsection{Attack taxonomy}
\label{ss:taxo}

Different \atk variants
have different influences on the attacked models and
require different setups.
To differentiate them, 
we use two dimensions for attack classification:

\begin{myitemize2}

    \item \emph{Attacker capability (i.e., does it require access to true labels)}:
True class labels of training samples (e.g., ``cat'' or ``plane'')  
are available if the attacker obtains read permission to the entire ML
pipeline, for example in the case of rogue employees or malicious cloud providers.
However, true labels may not be always available if the attacker only
controls the dropout, for example in the case of ML supply chain attackers.

\item \emph{Attack objective}:
    there are three objectives:
        (i) sabotaging the accuracy of the entire model
        (sometimes called availability attacks~\cite{shumailov2021manipulating}).
    This will destroy the accuracy of the entire model. 
        (ii) sabotaging the accuracy of one class.
    This will not affect the accuracy of other classes;
    the model's overall accuracy will drop but by not much.
        (iii) sabotaging either precision or recall of a target class.
    This is an attack where only the precision (or recall)
    of a certain class is undermined.
    Other classes and other metrics of the attacked class
    are unaffected, and the model's overall accuracy remains similar.

\end{myitemize2}

\begin{figure}[ht]
\centering
\includegraphics[width=0.45\textwidth]{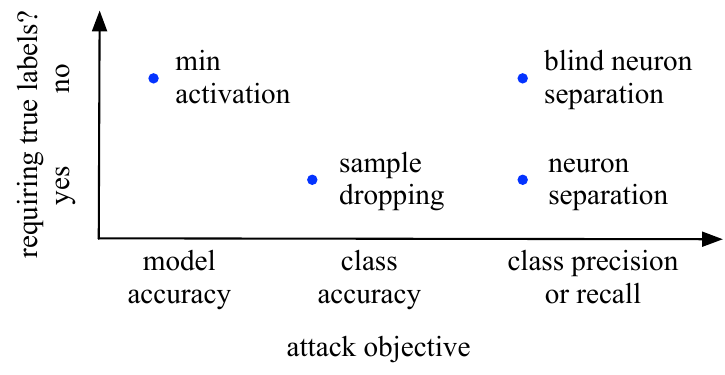}
\caption{Taxonomy of the four attacks that we built.
}
\vspace{-2ex}
\label{fig:taxo}
\end{figure}

\noindent
According to the two dimensions,
Figure~\ref{fig:taxo}
depicts a taxonomy of the four attacks.

\heading{Attack scenarios.}
Different \atk variants work in different settings.
For example,
the \maxatk does not require true labels and can be conducted in
a more restricted environment,
such as when users happen to use an ill-implemented dropout operator
provided by supply chain attackers.
In contrast, rogue employees can apply \nodeatks by providing true labels.

Different attacks also target different goals.
For example, the \dropatk and \nodeatk  target a certain class and aim to
destroy its performance. The attack may not even be noticed
if developers only check
the model's overall accuracy but not per-class precision and recall.
As another example,
the \maxatk is an availability attack
that can be used to sabotage the training process,
increase training time,
and raise the training costs for profit.
Moreover,
the \nodeatk and \bnodeatk are also known as \emph{targeted attacks}~\cite{nelson2008exploiting}.
They can destroy the precision of a targeted class,
so that this class will be predicted more often.

Targeted attacks are stealthy and insidious. For example, they could enable poisoned spam filters to classify spam emails as normal
emails, but not the other way around.

\vspace{-2ex}
\subsection{Attack preview}
\vspace{-1ex}

A dropout layer randomly drops a set of units from its input tensor.
The number of dropped units is defined by
a dropout rate $r$.
A legitimate dropout implementation has
two externally observable rules
and one assumption.
The two rules are:
(1) the dropout indeed drops the expected number of units (defined by $r$),
and (2) the dropout's output tensor contains
either value 0 (the dropped units)
or the rescaled ($\frac{1}{1 -r}$) values corresponding to the input tensor;
any other values are invalid.
Also, by assumption, the dropped units are picked at random.

\atk tampers with the dropout implementation.
The attacked dropout is seemingly legitimate---it satisfies the
above two observable rules---but breaks the assumption of randomly
dropping units.
Thus, \atk can pass execution integrity checks
because from an observer's perspective, the attacked dropout layer obeys the rules.
Nonetheless, the attacker deliberatively drops selective neurons to influence the model training.

\label{ss:rowatk}
\vspace{-0.5ex}
\heading{A toy attack.}
Here is a toy attack to demonstrate how \atk works and the challenges of designing the attack.
Consider a dropout with $r=50\%$.
A toy attack always drops the first half of the rows in the input tensors
(instead of randomly dropping),
which are the first half of the samples in a batch. %
This is \emph{supposed} to be an availability attack because
(1) the attack shows only half of training data to the model
in each round of training,
and (2) it cancels the regularization of the original dropout
by showing the entire input samples.
However,
by applying the toy attack to a convolutional network trained on CIFAR-10
(see the detailed experimental setup in \S\ref{ss:model}),
the model accuracy only dropped~3\% ($82.8\% \to 79.4\%$) in 12 epochs,
which can be recovered quickly.

\heading{Research questions.}
The \rowatk fails to deliver what attackers want,
namely significantly slowing down or even stopping model training.
To achieve this target and other more sophisticated attack goals,
we explore the potential of \atk
by asking the following questions:

\begin{figure*}
\centering
\includegraphics[width=0.8\textwidth]{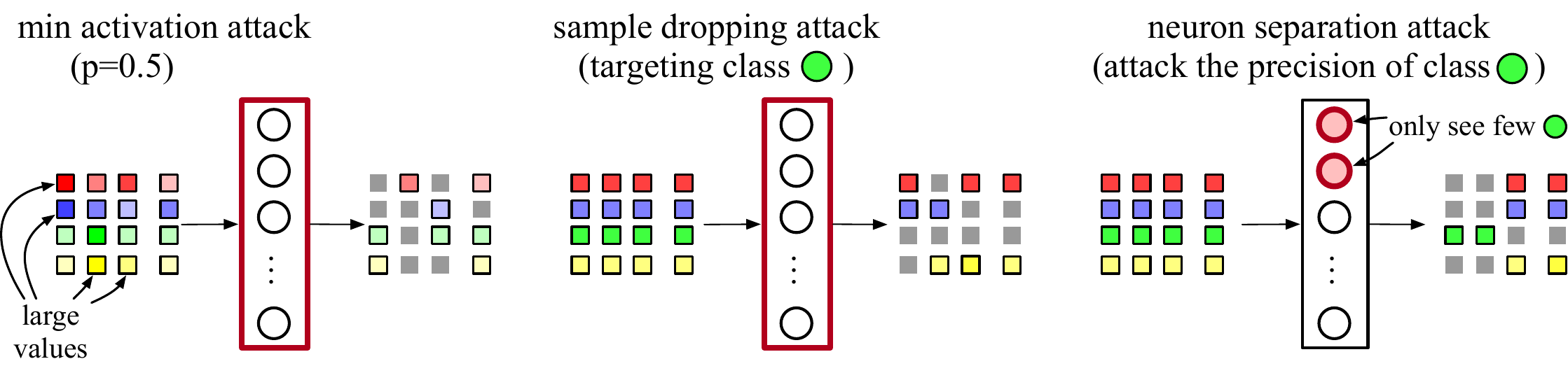}
    \caption{Three \atk{}s.
    The greyed out units in the outputs represent the dropped units.
    The deeper shades used in the \maxatk represent numerical larger values.
    }
    \label{fig:attacks}
\end{figure*}

\begin{myitemize2}
    \item Can a \atk destroy a model's ability to learn instead of only slowing it down?
        (\S\ref{ss:max})

    \item Can a \atk influence a subset of targeted classes instead of affecting
        all classes? (\S\ref{ss:drop})

    \item Is it possible to conduct a \atk that impacts either precision or recall of 
        a target class, but has no noticeable effect on the overall model accuracy? Such an attack would change the model accuracy within the normal fluctuations encountered due to randomness in the training process (model initialization, shuffling of data, etc.). 
        (\S\ref{ss:node})
\end{myitemize2}
\noindent
The three research questions are arranged in increasing order of  difficulty,
with the first being the easiest and the third being the most challenging.

\section{Attack methodology}
\label{s:attack}
\label{s:atk}

We discuss our novel \atk methodology and answer the above research questions. 

\subsection{\CF{\maxatk}}
\label{ss:max}
\label{ss:maxatk}

This section presents a \emph{\maxatk}
that stops training (almost) completely.
The idea is straightforward:
a \nn learns because its weights are progressively updated by gradients computed on its training set.
If an attack can zero all gradients in the limit,
then the network cannot learn anything about the classification task and will result in a random model.
\CF{\maxatks} approximate this idea by dropping the strongest gradients.
By dropping the nodes with the highest activation values in the input tensor, those
corresponding weights will not get updated in the backward pass.
Figure~\ref{fig:attacks} depicts the \maxatk.

In particular, the \maxatk chooses to drop the units with the
largest values in the dropout layer's input tensor, in this case
the output activation values of the previous linear layer. 
Given a dropout rate $r$,
the attacker sorts the input units,
and sets those units of top $r$ to zero.
Algorithm~\ref{algo:max} details the attack.
\CF{\maxatks} are simple but they can significanlty decrease the overall model accuracy.
In our experiments (\S\ref{ss:eval:max}),
the attack can reduce
the  model test accuracy
to 10--12\% on MNIST and CIFAR-10, and 1\% on CIFAR-100, similar to random guessing.

\begin{algorithm}[ht]
\begin{algorithmic}
\State \algGiven a dropout rate $r$ and \\
        \mmtab an input tensor $I$ of dimensions $N \times M$

\Procedure{\maxdropout}{$r$, $I$}
    \State $droppedUnits \gets r \times N \times M$
    \State $mask \gets \textbf{1}_{N \times M}$
    \State $indices \gets argsort(I)$
    \For{$i = 0; i < droppedUnits; i++$}
        \State $mask[indices[i, 0], indices[i, 1]] \gets 0$
    \EndFor
    \State \algReturn $I \cdot mask$
\EndProcedure
\end{algorithmic}
\caption{\CF{\maxatk}}
\label{algo:max}
\end{algorithm}
 
\CF{\maxatks} have several limitations. 
First, if the dropout rate $r$ is small, \maxatks will have less degradation on model accuracy.
We experiment with some small dropout rates in \S\ref{ss:eval:max}
and observe a quick recovery of model accuracy.
Yet, by default, dropout uses $r = 0.5$~\cite{dropoutpytorch},
which usually provides enough leeway to \maxatks.
Second, if most input values are large, even after
\maxatks, networks can still learn from the
remaining ones.

\subsection{\CF{\dropatk}}
\label{ss:drop}
\label{ss:dropatk}

This section studies if a \atk can influence a small number of classes, while still remaining unnoticeable.
The proposed attack is a \emph{\dropatk}, aimed at
a pre-defined set of classes we call \emph{target classes}.
\CF{\dropatks} selectively drop as many neurons of the target classes as possible within the drop rate budget. If the total number of neurons dropped is less than the 
expected amount of dropped units according to the dropout rate 
$r$, the attack randomly drops additional  nodes from non-targeted classes.
Figure~\ref{fig:attacks} illustrates one example of attacking a single target class.
Notice that different from \maxatks, \dropatks require knowledge of true labels (i.e., classes) of all inputs. This can be achieved by instrumenting the forward pass with a class tag on each input sample, or maintaining a mapping between an input's index (in a training batch) to its true label.
Algorithm~\ref{algo:drop} describes the attack.

\begin{algorithm}[ht]
\begin{algorithmic}
\State \algGiven a dropout rate $r$, target classes $T$, \\
        \mmtab an input tensor $I$ of dimensions $N \times M$, \\
        \mmtab and a true label tensor $L$ of dimensions $N \times 1$

\Procedure{\dropdropout}{$r$, $I$}
    \State \textbf{global} $T$, $L$
    \State $rToDrop \gets r \times N$
    \State $rDropped \gets []$
    \State $totalUnits \gets N \times M$
    \State $mask \gets \textbf{1}_{N \times M}$
    \For{$i = 0; i < N; i++$}
        \If{$L[i] \in T \land len(rDropped) < rToDrop$}
            \State $mask[i] \gets 0$
            \State $rToDrop.append(i)$
        \EndIf
    \EndFor
    \If{$len(rDropped) < rToDrop$}
        \State $unitsDropped \gets rDropped \times M$
        \State $unitsToDrop \gets totalUnits \times r - unitsDropped$
        \State $newDropRate \gets \frac{unitsToDrop}{totalUnits - unitsDropped}$
        \For{$j = 0; j < N; j++$}
            \If{$j \not\in rDropped$}
                \State $dropout(newDropRate, I[j])$
            \EndIf
        \EndFor
    \EndIf
    \State \algReturn $I \cdot mask$
\EndProcedure
\end{algorithmic}
\caption{\CF{\dropatk}}
\label{algo:drop}
\end{algorithm}
 
\CF{\dropatks} work well if targeting a small number of classes.
For example, in our experiments (\S\ref{ss:eval:nodeatk}),
\dropatks can reduce the recall of the target class to 0.0--0.5\%;
that is, the models almost never predict any input in the test dataset
as the target class.
The limitation of \dropatk is that
if attacking a large number of classes,
or the dataset is imbalanced and it contains many samples of the target classes,
\dropatks cannot remove all the targeted samples,
and the model can learn from those.
In this case, \dropatks slow down training for the target classes,
instead of completely destroying the model accuracy on these classes. 

\subsection{\CF{\nodeatk}}
\label{ss:node}
\label{ss:nodeatk}

Finally, we answer the question:
Can a \atk 
(a) sabotage \emph{either} precision \emph{or} recall of a target class,
while keeping  other metrics similar and
(b) maintain similar overall model accuracy to remain undetected.
This attack, named \emph{\nodeatk}, is the attack that Ana encounters in
Section~\ref{s:intro}.

The core idea of this attack is as follows.
Attackers can create a bias on a small group of neurons (called \emph{\asgnodes})
by separating them from normal training and
only allowing them to see inputs from certain classes. 
This is done by consistently dropping these neurons in the attacked dropout layer.
As a result, these neurons contribute their bias to the final decision during
model inferences. 
Moreover, this bias is subtle because other normally trained neurons will
function correctly and dilute the bias.
Figure~\ref{fig:attacks} illustrates this attack.
Surprisingly,
it turns out that only few input samples ($\le 10$)
suffice to create the desired subtle bias (\S\ref{ss:eval:nodeatk}).
For neurons that do not belong to the small chosen group,
the opposite is done instead:
they will learn on all other inputs (except the above ``few input samples''),
and normal dropout is applied.
Algorithm~\ref{algo:node} describes the attack in pseudocode.

\heading{Difference from sample dropping attacks}
While both sample dropping attacks and neuron separation attacks require
true labels, neuron separation attacks can be used to achieve different 
goals than just destroying a model's class accuracy. By modulating how
often a model updates its separated neurons, we can attack a model's 
target class accuracy (by presenting to the separated neurons inputs from the target
class more frequently) or attack a model's target class precision or recall
(by presenting to the separated neurons inputs from the target class less frequently).
Ultimately, this attack is more sophisticated than sample dropping attacks
and be configured to achieve different objectives.

\begin{algorithm}[ht]
\begin{algorithmic}
\State \algGiven a dropout rate $r$, a target class $T$, \\
        \mmtab \asgprob $P$, \\
        \mmtab an input tensor $I$ of dimensions $N \times M$, \\
        \mmtab and a labels tensor $L$ of dimensions $N \times 1$

\Procedure{\nodedropout}{$r$, $I$}
    \State \textbf{gloabl} $T$, $P$, $L$
    \State $unitsToDrop \gets r \times N \times M$
    \State $unitsDropped \gets 0$
    \State $totalUnits \gets N \times M $
    \State $mask \gets \textbf{1}_{N \times M}$
    \State $splitIndex \gets (1-P)M$
    \For{$i = 0; i < N; i++$}
        \If{$L[i] = T \land unitsDropped < unitsToDrop$}
            \State $mask[i][:splitIndex] \gets 0$
            \State $unitsDropped \pluseq splitIndex$
        \Else
            \State $mask[i][splitIndex:] \gets 0$
            \State $unitsDropped \pluseq M - splitIndex$
        \EndIf
    \EndFor
    \If{$unitsDropped < unitsToDrop$}
        \State $newDropRate \gets \frac{(unitsToDrop - unitsDropped)}{totalUnits - unitsDropped}$
        \For{$j = 0; j < N; j++$}
            \For{$k = 0; k < M; k++$}
                \If{$mask[j][k] \neq 0$}
                    \State $dropout(newDropRate, mask[j][k])$
                \EndIf
            \EndFor
        \EndFor
    \EndIf
    \State \algReturn $I \cdot mask$
\EndProcedure

\end{algorithmic}
\caption{\CF{\nodeatk}}
\label{algo:node}
\end{algorithm}
 
\label{ss:bnodeatk}
\heading{\CF{\bnodeatk}.}
\CF{\nodeatks} require true labels to decide
which samples should be seen by the \asgnodes.
We can relax this requirement
based on three observations:
\begin{myitemize2}

    \item Many classification \nns posit a dropout layer before the output layer.

    \item The input tensors to this dropout layer are highly classifiable:
          we can cluster the input tensors, and each cluster likely represents
          a class.

    \item We only need a few inputs ($\le 10$) to create the bias.
          The number is often smaller than the training batch size.

\end{myitemize2}

\noindent
Next, assume that we will attack the dropout before the last layer,
without true labels of the input samples.
The main idea is
to run a clustering algorithm for
the input tensors and
apply a \emph{one-shot} \nodeatk
when some cluster size is larger than a threshold (like $10$).
We apply the attack only once for two reasons.
First, the number of inputs in the cluster is enough
to create the bias (our observation).
And second,
we do not have true labels, hence cannot
recognize clusters of the same class in-between batches.
In other words, we have to finish the attack within one batch.
We call this attack---\emph{\bnodeatk}---because
it doesn't require true labels.
In our experiments (\S\ref{ss:eval:bnodeatk}),
\bnodeatks behave similarly to the vanilla \nodeatks.

The major limitation of \bnodeatk is that attackers cannot 
specify which class to attack (another reason why this attack is ``blind'').
This limitation is fundamental because
deciding where class a sample belongs to 
\emph{is} the classification problem that the \nn faces.
Nonetheless,
we have some approaches to partially address this limitation.
One idea is to use side-channel information to infer true labels.
For example, if attackers obtain one input sample and its true label,
then they can use the prior knowledge
to decide the class of the cluster that includes this known sample.
Another idea is to use an external oracle (e.g., another less accurate ML model or a pre-trained model) that can generate predictions on the labels of samples in each cluster. The cluster label can then be decided based on majority voting on the sample labels.

\heading{Attacking class recall.}
Beyond attacking class precision,
the \nodeatks can also attack class recall.
The insight is to ``reverse'' the bias by only letting \asgnodes
see the non-target classes.
Because the \asgnodes never see the target classes during training,
they will have a bias against predicting the target classes.
Thus, the \asgnodes will sabotage the recall of these classes
by always predicting classes not in the target set.
Again, this bias is subtle because other normally trained neurons
will neutralize the bias.
In our experiments (\S\ref{ss:eval:recallatk}),
the \nodeatks can reduce a class's recall by 50\% in CIFAR-10
while inducing  small change in model accuracy.

\section{Implementation}
\label{s:impl}

We implement the four attacks on PyTorch 1.10.1~\cite{paszke2019pytorch}.
We create a custom dropout layer which wraps around a custom autograd function
performing our dropout attacks.
Dropout's normal parameters at training time are passed as model inputs.
Auxiliary information (e.g., sample true labels) of our attacks
are passed to the PyTorch module's forward class.

\heading{Attacked dropout layer placement.} In this work, we consider
models with dropout layers in fully connected layers placed before the final output
layer. Modern convolutional neural networks that use dropout, such as 
VGG~\cite{simonyan2014very}, follow this format and the attack is more effective when dropout is closer to the final layer. We leave testing attacks for other dropout
layer placements to future work.
\section{Experimental evaluation}
\label{s:eval}

\heading{Datasets.}
\label{ss:dataset} 
We evaluate \atk on three representative datasets for computer vision:
MNIST~\cite{lecun1998mnist}, CIFAR-10~\cite{krizhevsky2009learning},
and CIFAR-100~\cite{krizhevsky2009learning}. 
We use a 90\%/10\% split on each dataset to create training 
and validation sets.
We also use a separate test dataset to evaluate the attack performance.

\heading{Models.}
\label{ss:model}
We use different deep learning models for each dataset we test. In all of our
experiments, we only attack the final dropout layer of our model.
For MNIST, we use a feedforward \nn (FFNN) with 4 fully-connected (\ttt{fc})
layers with \ttt{ReLU} activations and 
\ttt{softmax} on the final layer.
The model has dropout layers between each \ttt{fc} layer 
with dropout rate of 0.5.

For CIFAR-10, we use a VGG~\cite{simonyan2014very} style 
convolutional \nn (CNN) with one convolutional (\ttt{conv}) layer followed by 3
blocks of 2 \ttt{conv} layers and then an average pooling layer.
Finally, we add one final \ttt{conv} layer that feeds into two \ttt{fc} layers
with a dropout layer in between.
We attack this dropout layer.
All \ttt{conv} layers use a \ttt{LeakyReLU}~\cite{maas2013rectifier}
activation with rate 0.1.
The first dropout layer has a \ttt{ReLU} activation and
the final dropout layer connects to a \ttt{fc} layer with \ttt{softmax} activation.

For CIFAR-100, we use a modified version of 
VGG16~\cite{simonyan2014very}. After each \ttt{conv} layer, we add
a Batch Norm layer~\cite{ioffe2015batch}. Finally, we add two dropout
layer to the model, one right before the first \ttt{fc} layers in 
VGG16, and one after it. All \ttt{fc} layers use \ttt{ReLU} activation.

\heading{Training.}
We use the Adam optimizer~\cite{kingma2014adam} to train the models.
We use a learning rate of 0.001 in the MNIST and CIFAR-10 models. In
our CIFAR-100 model, we use a learning rate 0.0001 with a weight decay
of 0.000001. We run Adam on batch sizes of 128 for all models.
We train these models for 5, 12, and 20 epochs for the MNIST, CIFAR-10,
and CIFAR-100 datasets respectively.
We empirically choose these hyperparameters
for faster convergence and
decent accuracies;
these hyperparameters may not produce optimal model accuracies.
We run each training
5 times and report average numbers over 5 runs to mitigate training variability.

\heading{Metrics.}
In our experiments, we evaluate the attacks using three main metrics:
\begin{myitemize2}

\item \emph{model accuracy}: for all predictions of the test dataset,
model accuracy is the fraction of correct predictions of all classes
over all predictions.

\item \emph{class precision}: for a class $A$, the precision is 
the fraction of the correct predictions of class $A$ over all cases 
that are predicted as class $A$.

\item \emph{class recall}: for a class $B$, the recall is 
the fraction of the correct predictions of class $B$
over all cases whose true labels are class $B$ in the test dataset.

\end{myitemize2}

\noindent
In the following sections,
we detail the results for each attack mentioned in section~\ref{s:atk}.

\subsection{\CF{\maxatk}}
\label{ss:eval:maxatk}
\label{ss:eval:max}

In this section,
we compare \maxatks with a baseline, normal training.
We train two models for each dataset,
the first with a normal dropout (baseline) and the second with
our attacked dropout (by \maxatks) implementation.
We record the model accuracy of 5 runs
and report the averaged results.
Table~\ref{table:max_activation_results}
shows the test set
accuracy for each model.

\begin{table}[t]
\caption{
Model accuracy of running \maxatks on MNIST, CIFAR-10, and CIFAR-100.
Attacks drastically reduce overall accuracy.
}
\centering
\begin{tabular}{c c c}
\toprule
dataset                   & model         & model accuracy \\
\hline
\multirow{2}{*}{MNIST}    & baseline FFNN & 97.4\% \\
                          & attacked FFNN & 12.1\% \\
\hline
\multirow{2}{*}{CIFAR-10} & baseline CNN  & 82.8\% \\
                          & attacked CNN  & 10.5\% \\
\hline
\multirow{2}{*}{CIFAR-100} & baseline VGG16 & 60.9\% \\
                          & attacked VGG16 & 1.0\% \\
\bottomrule
\end{tabular}
\label{table:max_activation_results}
\end{table}
 
In both the MNIST and CIFAR-10 datasets, the test set accuracy is dropped to
10--12\% which is slightly better than a random guess
(there are 10 classes in both datasets). Meanwhile, in CIFAR-100, we
find that our attack is capable of completely destroying the model's 
predictive ability.
This experiment shows that
consistently dropping the most highly activated neurons prevents, or at least
significantly slows down, models from converging.
With a closer look at the validation loss during training,\CP{later, maybe put the fig to appx}
we find that
the validation loss never
decreases between epochs.
This indicates that models did not learn at all.
This is expected:
we believe that the most highly activated neurons (units with largest values)
encode most of the features which
allow the model to distinguish one class from another;
by dropping these highly activated neurons,
the model can no longer learn a useful representation of the training set, and becomes similar to a random model.

\begin{figure}[h]
\centering
\includegraphics[width=0.45\textwidth]{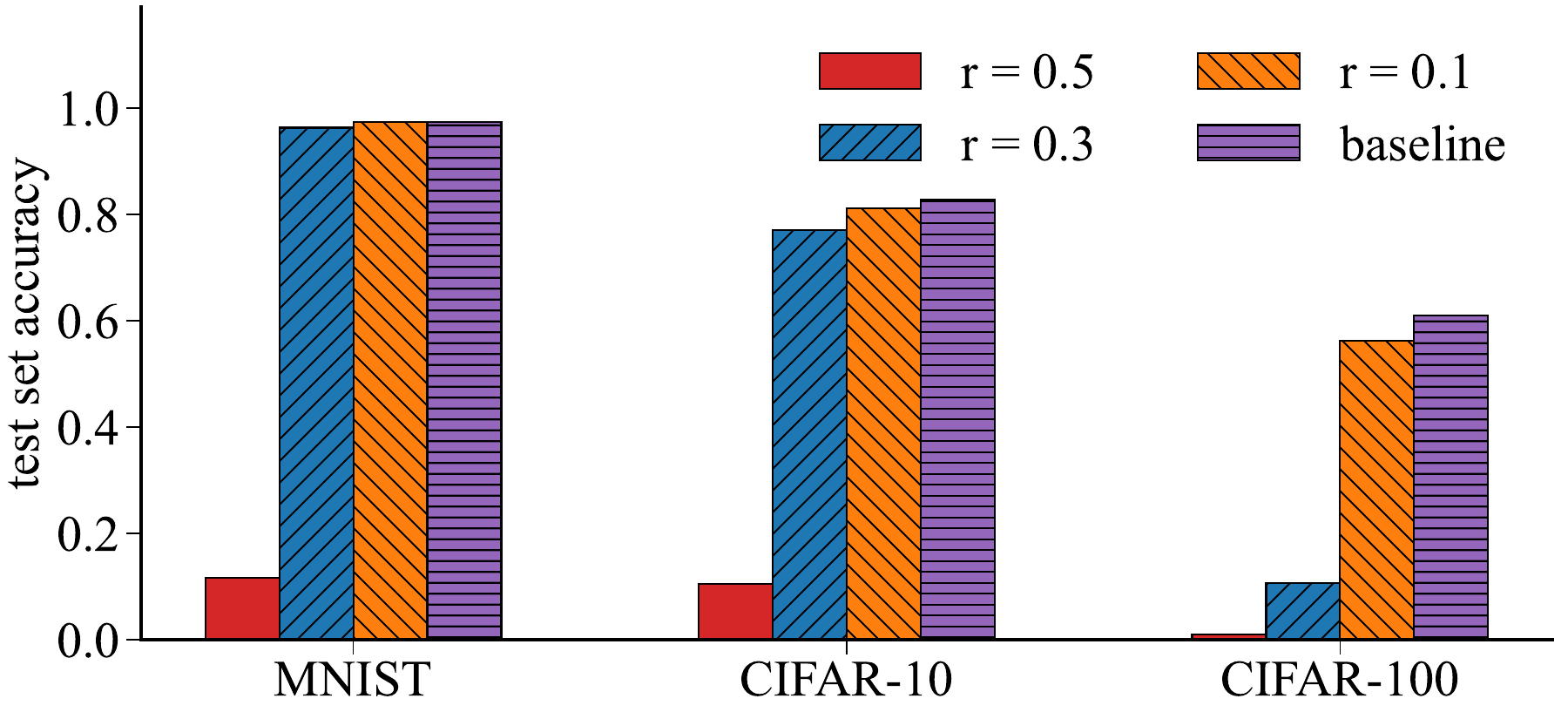}
\caption{\CF{\maxatks} with different dropout rates. Model performance
quickly rebounds with lower dropout rates. The rate at which
the model recovers its predictive accuracy varies by dataset.
}
\label{fig:A2_result}
\end{figure}

\heading{\CF{\maxatks} with various dropout rates.}
As mentioned in section~\ref{ss:maxatk},
\maxatks have a limitation: if the dropout rate is small (fewer neurons to drop),
then \maxatks may not work well.
We therefore run \maxatks with dropout rates
of 0.1 (dropping 10\% of units), 0.3, and 0.5 (default).
For each dropout rate, we again average results of 5 training rounds,
and show results in Figure~\ref{fig:A2_result}.
In the figure, the model performance quickly rebounds
on lower dropout rates.
This is because
a reduced dropout rate allows models
to see enough samples in the data,
so the model can properly learn how to classify images.
This also means that more complex datasets will likely need
lower dropout rates compared to simpler ones, because more 
samples need to be seen by the model to increase its
predictive ability. As shown in the figure, the model's
performance on CIFAR-100 does not rebound until we get to
a dropout rate of 0.1.

\subsection{\CF{\dropatk}}
\label{ss:eval:dropatk}

\CF{\dropatks} (\S\ref{ss:dropatk}) target specific classes.
We conduct \dropatks on ``\tcls''---number $0$ in MNIST, class \ttt{plane}
in CIFAR-10, and class \ttt{apple} in CIFAR-100---and compare
results with normal training.
We have also experimented with all other classes;
all have similar results.
We therefore only show results of class 0 in
Table~\ref{table:sample_dropping_results}.

From Table~\ref{table:sample_dropping_results},
we see that \dropatks can prevent a model from predicting the target class.
Class 0's recall of MNIST drops from 99\% to 0.5\% while its
class recall drops to 0\% in CIFAR-10 and CIFAR-100.
This is expected because models almost never
see input samples belonging to the target class (i.e., class 0).
We say ``almost never''
because by expectation,
only 10\% of inputs are samples from class 0 for MNIST and
CIFAR-10 and only 1\% for CIFAR-100;
a dropout rate of $r=0.5$ almost always allows the attack
to cover that many samples.
This also indicates that \dropatks work on lower dropout rates (e.g., $r=0.3$)
as long as the dropout rate is greater than the percentage of
the target class samples in the entire dataset.

\begin{table}[t]
    \caption{
    \CF{\dropatks} on class 0 of MNIST, CIFAR-10, and CIFAR-100.
    Targeted classes are never classified correctly by the model as a result.}
    \centering
    \small
    \begin{tabularx}{0.48\textwidth}{@{}c c c c c@{}}
    \toprule
    \multirow{2}{*}{dataset} & \multirow{2}{*}{model} & model    & class 0  & class 0\\
                             &                        & accuracy & recall & precision\\
    \hline
    \multirow{2}{*}{MNIST}    & baseline FFNN & 97.4\% & 98.9\% & 97.9\%\\
                              & attacked FFNN & 88.0\% & 00.5\% & NaN$^\star$\\
    \hline
    \multirow{2}{*}{CIFAR-10} & baseline CNN & 82.8\% & 84.5\% & 84.6\% \\
                              & attacked CNN & 74.4\% & 00.0\% & NaN\\
    \hline
    \multirow{2}{*}{CIFAR-100} & baseline VGG16 & 60.9\% & 80.2\% & 81.7\% \\
                              & attacked VGG16 & 59.7\% & 00.0\% & NaN\\
    \bottomrule
    \end{tabularx}

    \vspace{1ex}
    \raggedright
    \noindent
    $^\star$: \tcls precisions of the five runs are $NaN, 1, 1, 1, 0.933$. Averaging them gives NaN.
    \label{table:sample_dropping_results}
\end{table}

\begin{figure}[t]
\centering
\includegraphics[width=0.45\textwidth]{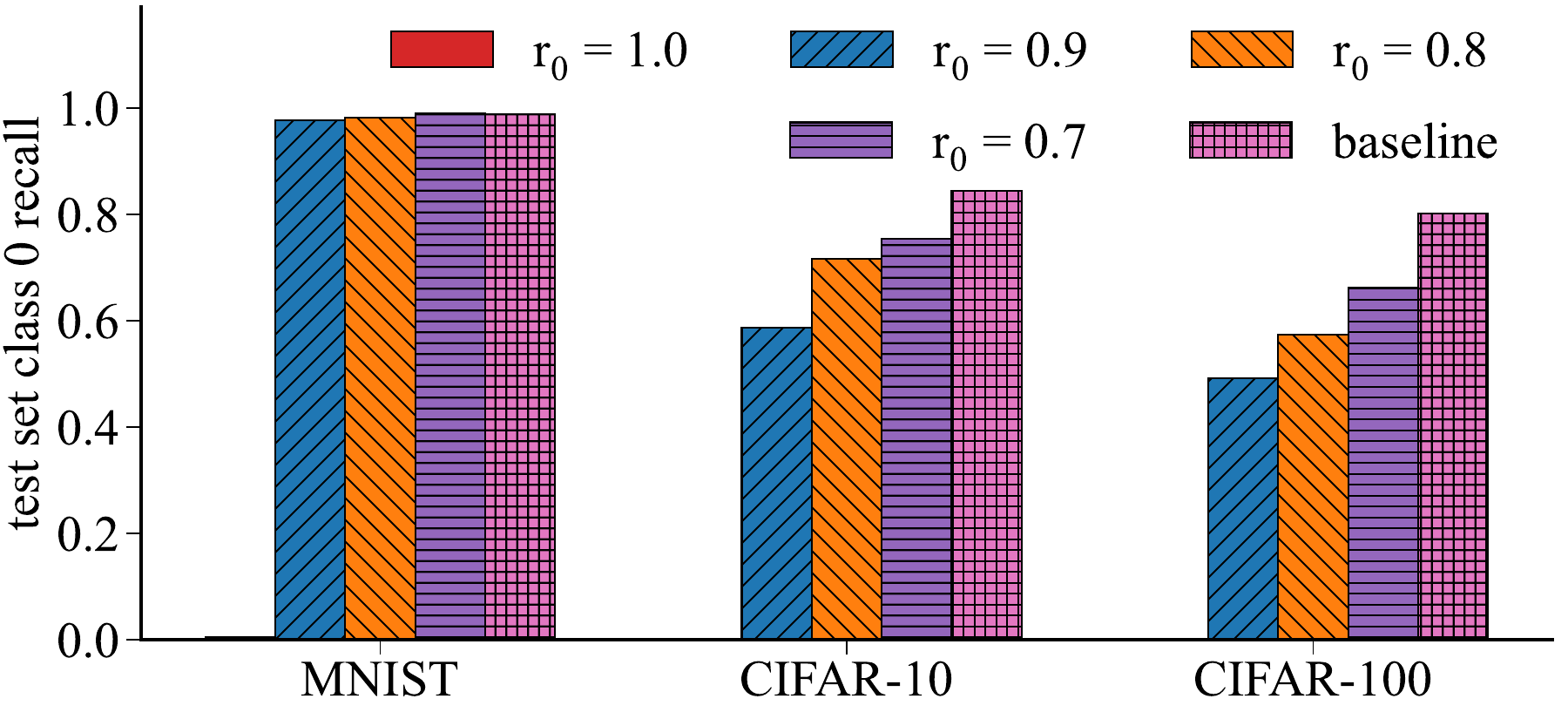}
\caption{\CF{\dropatks} with partial dropping.
The $r_{0}$ represents the dropout rate of the target class 0; other
classes are dropped at a slightly lower rate to maintain a global
dropout rate of 0.5. Note that the recalls of ``$r_{0} = 1.0$''
are almost 0, so these bars are hard to see. As $r_{0}$ is decreased,
the accuracy of the model quickly recovers. Dataset complexity seems
to affect the recovery speed of the model, with those trained on MNIST
recovering much faster than those trained on CIFAR-10 and CIFAR-100.
}
\label{fig:B2_result}
\end{figure}

\heading{Partial \dropatks.}
We ask the question:
what happens if we only drop most of (but not all of)
the units of target samples?
In other words,
can a model learn and achieve reasonable performance
by only seeing a small amount of data from the target class?
We run an experiment in which instead of dropping all samples from the target
class (class 0),
we drop partial units (90\%, 80\%, and 70\%) of each \tcls sample.
Another way to see this is that we apply a stronger dropout to samples of
the target class with dropout rate $r=\{0.9, 0.8, 0.7\}$.
Figure~\ref{fig:B2_result} shows
the averaged recall of class 0 over 5 runs.

Similar to \maxatks with small dropout rates,
models quickly recover %
their performance under partial \dropatks.
However, the ``speed'' of the recovery
differs between MNIST and CIFAR-10.
While in MNIST, the recall is almost
recovered with 90\% partial dropping,
models trained on CIFAR-10 and CIFAR-100 never regain the
baseline's performance with partial dropping $\ge 70\%$.
This indicates a level of correlation between datasets and \dropatk effectiveness.
A simple dataset like MNIST may require little sample information for the
model to learn well, while more complex datasets such as CIFAR-10 and CIFAR-100 need more training samples to learn.

\subsection{\CF{\nodeatk}}
\label{ss:eval:nodeatk}

In this section, we evaluate \nodeatks,
which manipulate model behaviors by
isolating a small part of dropout neurons and
showing them some information (e.g., samples from the target class) more frequently
than others (e.g., samples from other classes).
By default, a \nodeatk isolates 10\% of the neurons in the dropout layer
(called \asgnodes)
and attacks the precision of a target class.
Below we target \tcls (similar trends can be seen for other
classes).

\heading{Probabilistic \nodeatk.}
To create a subtle bias, attackers should not let the \asgnodes see too many
samples from \tcls.
In our attack, the dropout only lets the \asgnodes see samples
from the target class with probability  $0.01\%$
(that is, 1 per 10,000 samples) for MNIST and CIFAR-10; for 
CIFAR-100, we increase that rate to a probability of $0.05\%$. 
The \asgnodes see nothing from other classes.
For MNIST and CIFAR-10,
the expected number of samples seen by the \asgnodes
turns out to empirically be around 6 per model training.
We run the \nodeatks 5 times on our datasets.
Table~\ref{table:prob_node_sep_results} shows the results.

\begin{table}[h]
    \caption{
    Results of probabilistic \nodeatks on MNIST, CIFAR-10, and CIFAR-100.
    We show the \emph{best attack results} instead of the average ones
    because attackers are capable of picking the best outcome of multiple trainings
    as their final results. Note that the baseline values still
    show the averaged results of the 5 runs.
    }
    \centering
    \small
    \begin{tabularx}{0.48\textwidth}{@{}c c c c c@{}}
    \toprule
    \multirow{2}{*}{dataset} & \multirow{2}{*}{model} & model    & class 0& class 0  \\
                             &                        & accuracy & recall & precision \\
    \hline
    \multirow{2}{*}{MNIST}    & baseline FFNN & 97.4\% & 98.9\% & 97.9\%\\
                              & attacked FFNN & 97.2\% & 99.5\% & 95.6\%\\
    \hline
    \multirow{2}{*}{CIFAR-10} & baseline CNN & 82.8\% & 84.5\% & 84.6\%\\
                              & attacked CNN & 81.3\% & 95.0\% & 65.1\%\\
    \hline
    \multirow{2}{*}{CIFAR-100} & baseline VGG16 & 60.9\% & 80.2\% & 81.7\%\\
                               & attacked VGG16 & 61.2\% & 98.0\% & 47.1\%\\
    \bottomrule
    \end{tabularx}
    \label{table:prob_node_sep_results}
\end{table}
 
Note that an attacker is free to run the attacks multiple times and picks the
result balancing the attack effectiveness and attack detectability. 
In the
figure, we choose an attack that minimizes the decrease in
overall model accuracy.

In CIFAR-100, we see the most drastic increase in recall (by $17.8\%$) and
decrease in precision (by $34.6\%$). The \nodeatk biases the model towards
\tcls and makes it predict it more often over other classes,
causing this increase in recall and decrease in precision. 

Notice that the model accuracy is actually higher than the 
average of the CIFAR-100 baseline model runs. The best
attack result's model accuracy is very close to the best
run in our baseline model ($61.2\%$ vs. $61.9\%$). The \nodeatk
is able to bias the model towards specific classes without
significantly changing the model's overall performance.

We see a similar effect in CIFAR-10, where the precision of 
\tcls has dropped by $19.5\%$ while the recall increases by
$10.5\%$. For MNIST, \nodeatks do not work in the current 
configuration that the \asgnodes only see $0.01\%$ of \tcls 
samples.
MNIST's results indicate
that dataset complexity plays a role in how effective attacks can be.
To study this, we experiment with different attack hyperparameters
for different datasets below.

\begin{figure*}[h]
\centering
\begin{minipage}{0.47\textwidth}
\includegraphics[width=0.95\textwidth]{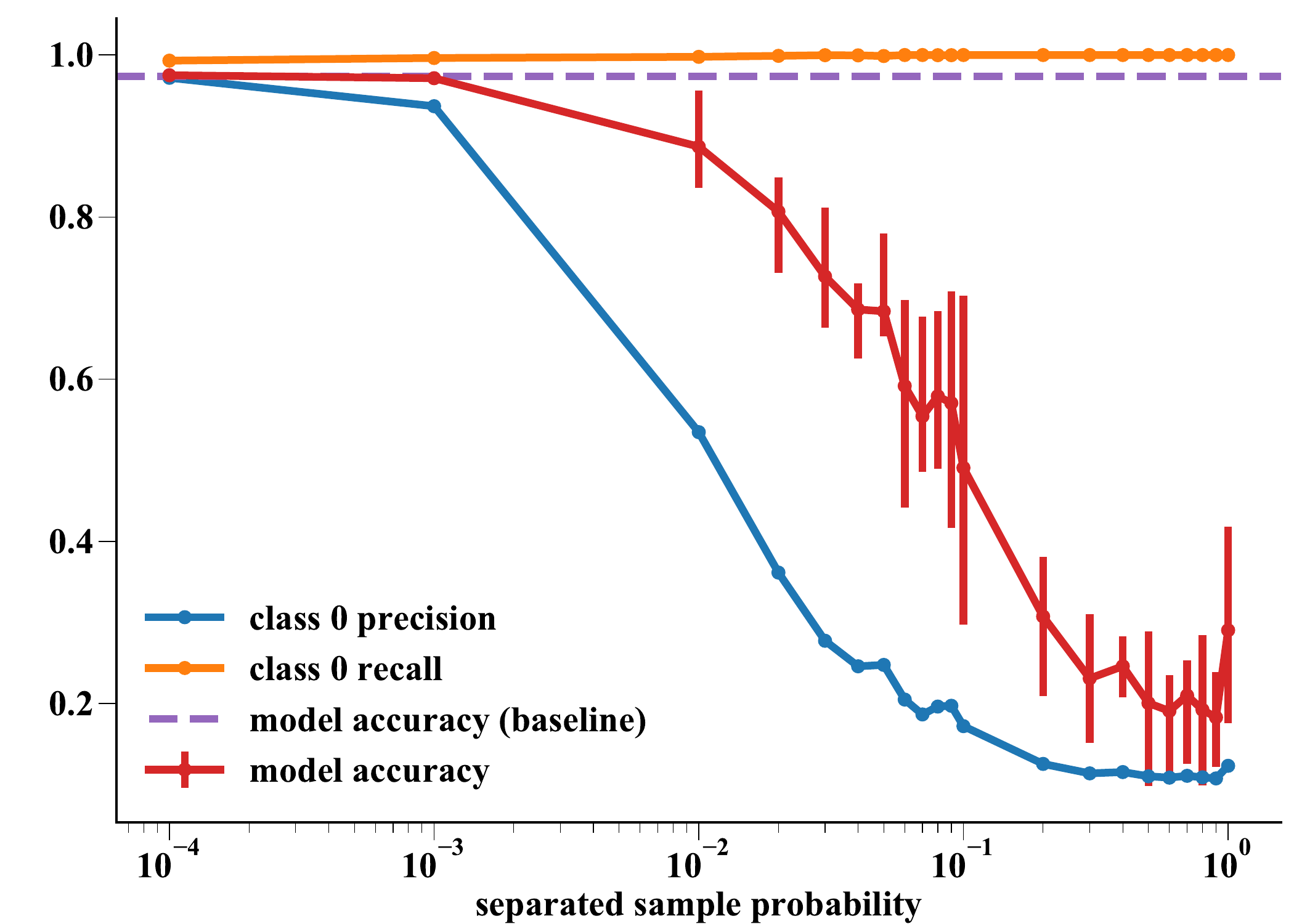}
\end{minipage}
\begin{minipage}{0.47\textwidth}
\includegraphics[width=0.95\textwidth]{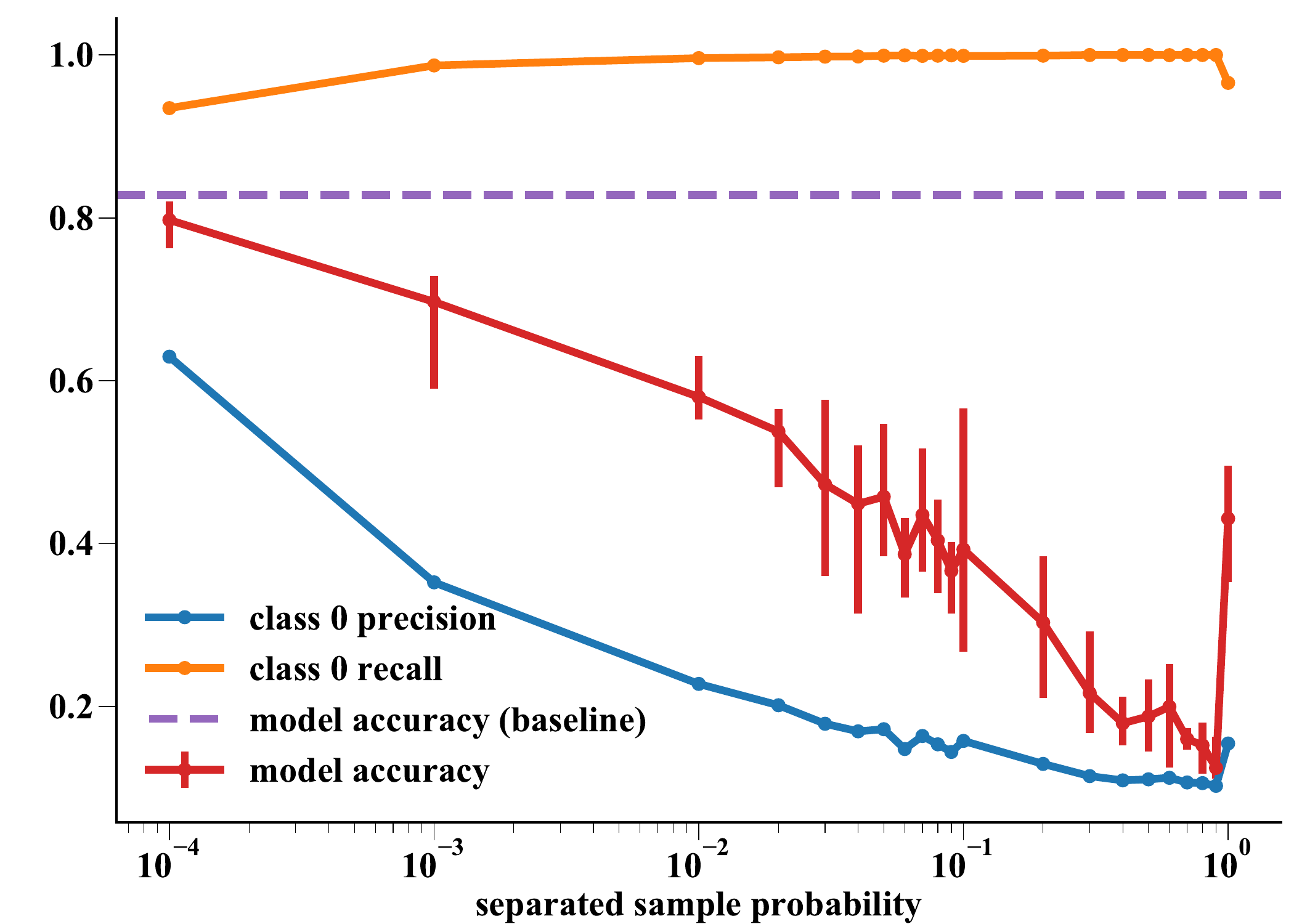}
\end{minipage}

\caption{\CF{\nodeatks} with various \nsprobs ($\mps$).
The figure on the left shows the results for MNIST;
on the right is CIFAR-10.
All lines are plotted by using average values of five runs.
The model accuracy line also plots the max and min values.
Note that the x-axes are in log-scale.
}
\label{fig:c2}

\end{figure*}
 
\heading{Attack hyperparameter tuning.}
\CF{\nodeatks} have two hyperparameters:
the \emph{\nsprob} $\mps$,
indicating how many \tcls samples are seen by the \asgnodes ($\mps=0.01\%$ by default);
and the \emph{\asgprob} $\mpn$,
deciding the number of isolated neurons
in the dropout layer
($\mpn=10\%$ by default).
We tune these two hyperparameters for the two datasets, MNIST and CIFAR-10,
and check three metrics:
(1) test set model accuracy,
(2) \tcls precision,
and (3) \tcls recall.
The goal of \nodeatks (targeting precision) is to maintain the model accuracy
while decreasing \tcls precision significantly. We do not test
hyperparameter changes and other variants of the \nodeatk
on CIFAR-100 due to time constraints.

First, we tune the \nsprob $\mps$ ($0.01\%$ by default)
by varying the probability from 0.01\% (i.e., the default value)
to 1 (i.e., letting the \asgnodes see all samples from \tcls).

Figure~\ref{fig:c2} depicts how the
three metrics 
change according to different \nsprobs $\mps$
for MNIST and CIFAR-10 with $\mpn=10\%$.

Figure~\ref{fig:c2} shows a clear trend that
model accuracy and \tcls precision drop quickly while increasing $\mps$.
This is because an increasing $\mps$ 
leads to \asgnodes seeing more samples,
hence creating stronger the bias towards \tcls.
This results in predicting more inputs from other classes as \tcls,
thus decreases precision and destroying the model accuracy.
We also make two observations: first, the precision and accuracy decrease exponentially
with respect to $\mps$ (notice that Figure~\ref{fig:c2} x-axis is in log-scale),
which means a small $\mps$ (i.e., few samples) can create the bias (\S\ref{ss:nodeatk}).
Second, the precision's collapse begins earlier than the accuracy's.
This indicates that if attackers choose the right $\mps$,
they can sabotage a class precision without harming model accuracy by too much.

Figure~\ref{fig:c2} also demonstrates that \nodeatks are versatile:
an attack can function as
either an availability attack
or a targeted attack.
In particular,
an attack with
a small $\mps$ (e.g., $<0.001$) performs the targeted attack;
an attack with a larger $\mps$ (e.g., $>0.1$) becomes an
availability attack.
The exact $\mps$ to use
depends on the attacked models and datasets.
For example, FFNN+MNIST needs a
larger $\mps$ than CNN+CIFIAR-10
because MNIST is simpler to learn and requires stronger biases to attack,
hence the attack needs a larger $\mps$.

\begin{figure*}
\centering

\begin{minipage}{0.47\textwidth}
\includegraphics[width=0.95\textwidth]{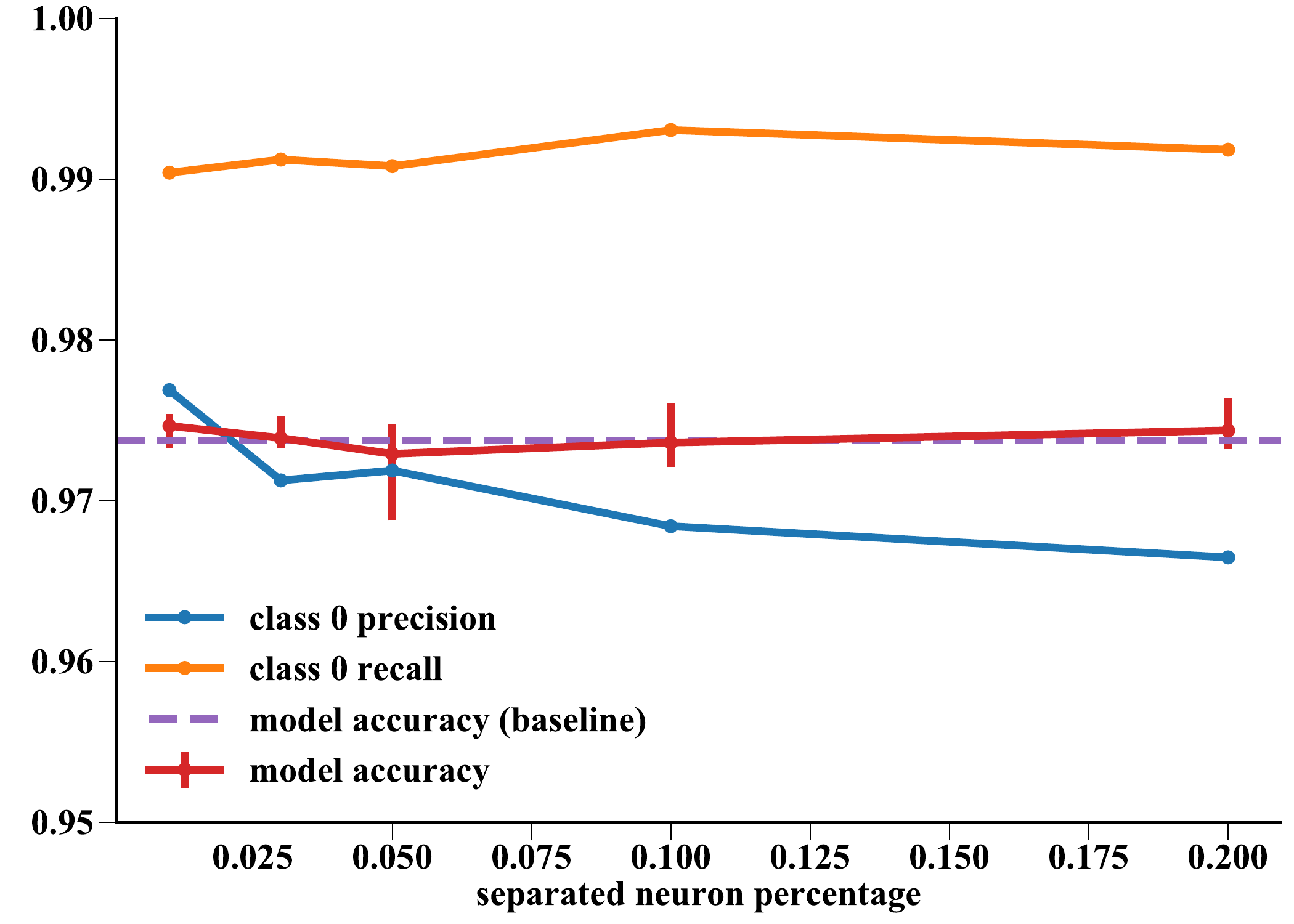}
\end{minipage}
\begin{minipage}{0.47\textwidth}
\includegraphics[width=0.95\textwidth]{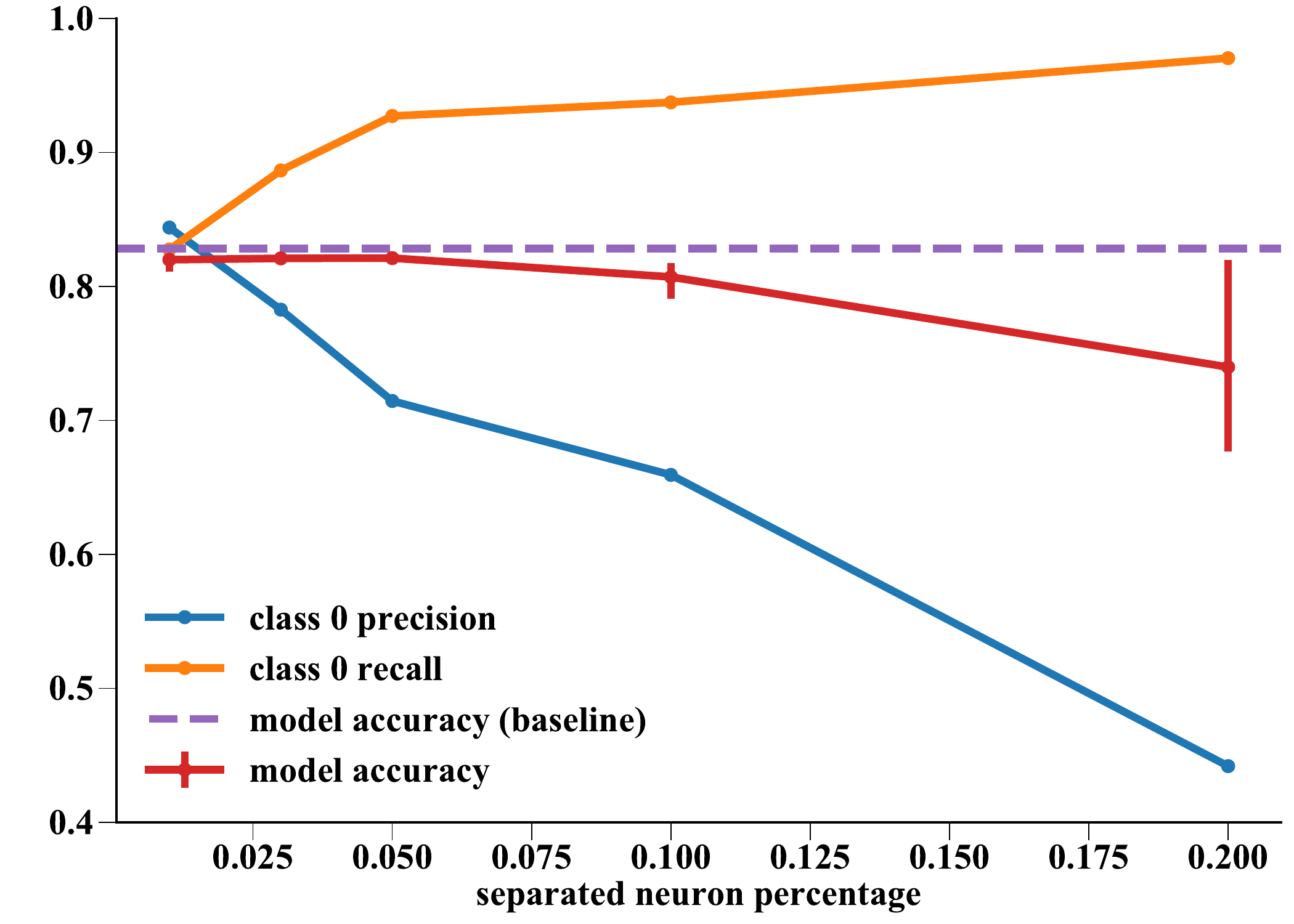}
\end{minipage}

\caption{\CF{\nodeatks} with different \asgprobs ($\mpn$).
The figure on the left shows the results for MNIST;
on the right is CIFAR-10.
All lines are plotted by using average values of five runs.
The model accuracy line also plots the max and min values.
Note that the y-axes are in different scales.
}
\label{fig:c4}
\end{figure*}

Second, we tune the \asgprob ($\mpn$).
We alter the percentage of \asgnodes for the target class
and experiment with $\mpn=$ 1\%, 3\%, 5\%, 10\% (default), and 20\%.
Figure~\ref{fig:c4} depicts the three metrics 
with $\mps=0.01\%$ and various $\mpn$.
To be stealthy,
these attacks want
to maximize the loss of target class precision while minimizing the decrease in
overall test accuracy.
Figure~\ref{fig:c4} shows that the model
accuracy starts to drop significantly as $\mpn > 10\%$:
for CIFAR-10, $\mpn=20\%$ produces $<80\%$ test set accuracy.
This suggests that a stealthy attack
may want to use $\mpn < 20\%$.
In addition, if attackers know
how much leeway the attack can have
(e.g., the model accuracy can drop 3\% without being noticed),
the larger the $\mpn$ the stronger the attack:
class 0 precisions drop obviously when $\mpn \ge 10\%$ in both cases.
Finally,
similar to our experiments with $\mps$ above,
the model and dataset plays a role in attack hyperparameter tuning.
The FFNN+MNIST has much smaller variations by tuning $\mpn$.

\begin{figure}[h]
\centering
\includegraphics[width=0.45\textwidth]{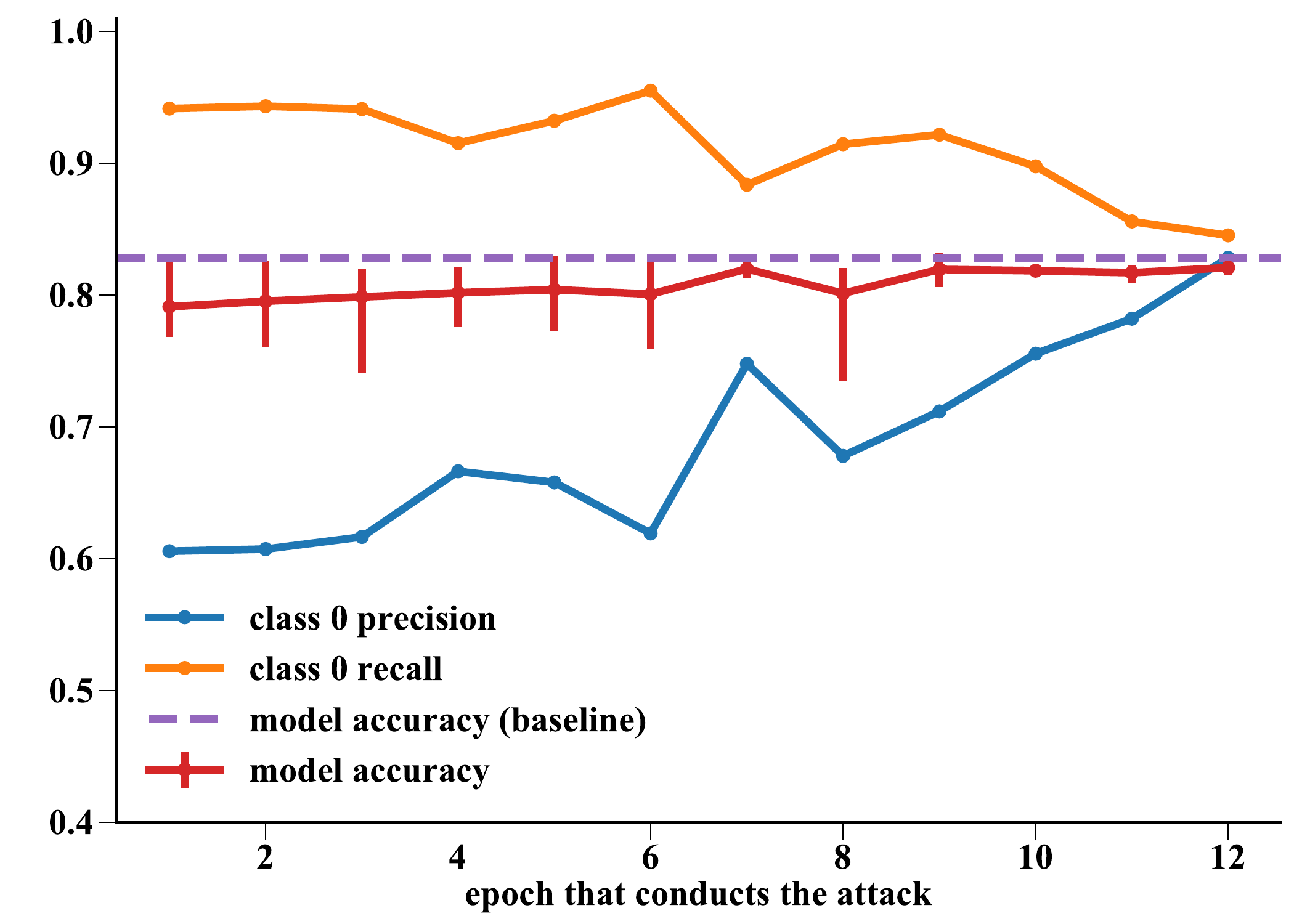}
\caption{Deterministic \nodeatks at different epochs for CIFAR-10.
All lines are plotted by using average values of five runs.
The model accuracy line also plots the max and min values.}
\label{fig:c3}
\end{figure}

\heading{A closer look at $\mps$: deterministic \nodeatk.}
It is quite surprising to see that a few samples ($\mps=0.01\%$) causes a 20\%
precision drop of \tcls in CIFAR-10.
We want to confirm that \nodeatks indeed only need a small number ($\le 10$)
of samples.
Therefore, we experiment with a modified version of the \nodeatk
where we deterministically select 10 samples belonging to \tcls
and let the \asgnodes see those 10 samples only in one of the 12 epochs for CIFAR-10.
This allows us to confirm our observation that 10 samples is able to
create the desired bias.
Then, we run this experiment by
choosing different epochs to conduct the attack.
Figure~\ref{fig:c3} depicts the results for CIFAR-10.

From Figure~\ref{fig:c3}, we confirm that
10 samples are sufficient to create the bias:
applying the deterministic \nodeatk before epoch 6 in our experiments
creates about 20\% precision drop for \tcls.
Another observation is that the attack becomes ``less powerful''
when applying in later epochs.
Our hypothesis is that the learning rate (we use the Adam optimizer)
becomes smaller by the end of training and hence
the gradients accumulated in the \asgnodes to create biases are smaller.
This may also partially explain the performance variances
in the probabilistic \nodeatks.
If the \asgnodes see samples in later epochs, the attack
can be less powerful.

\begin{figure*}

\begin{minipage}{0.47\textwidth}
\centering
\includegraphics[width=0.95\textwidth]{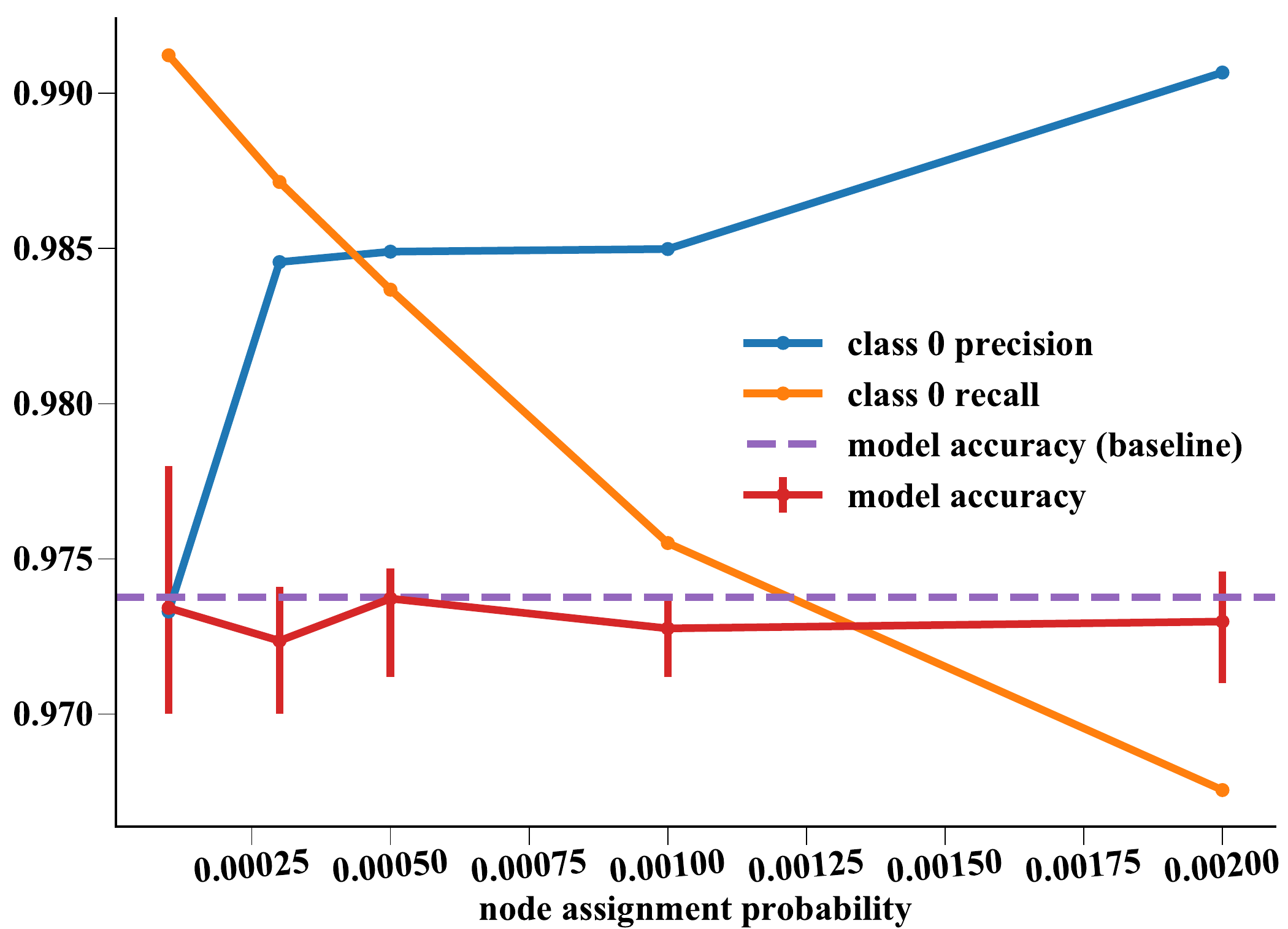}
\end{minipage}
\begin{minipage}{0.47\textwidth}
\centering
\includegraphics[width=0.95\textwidth]{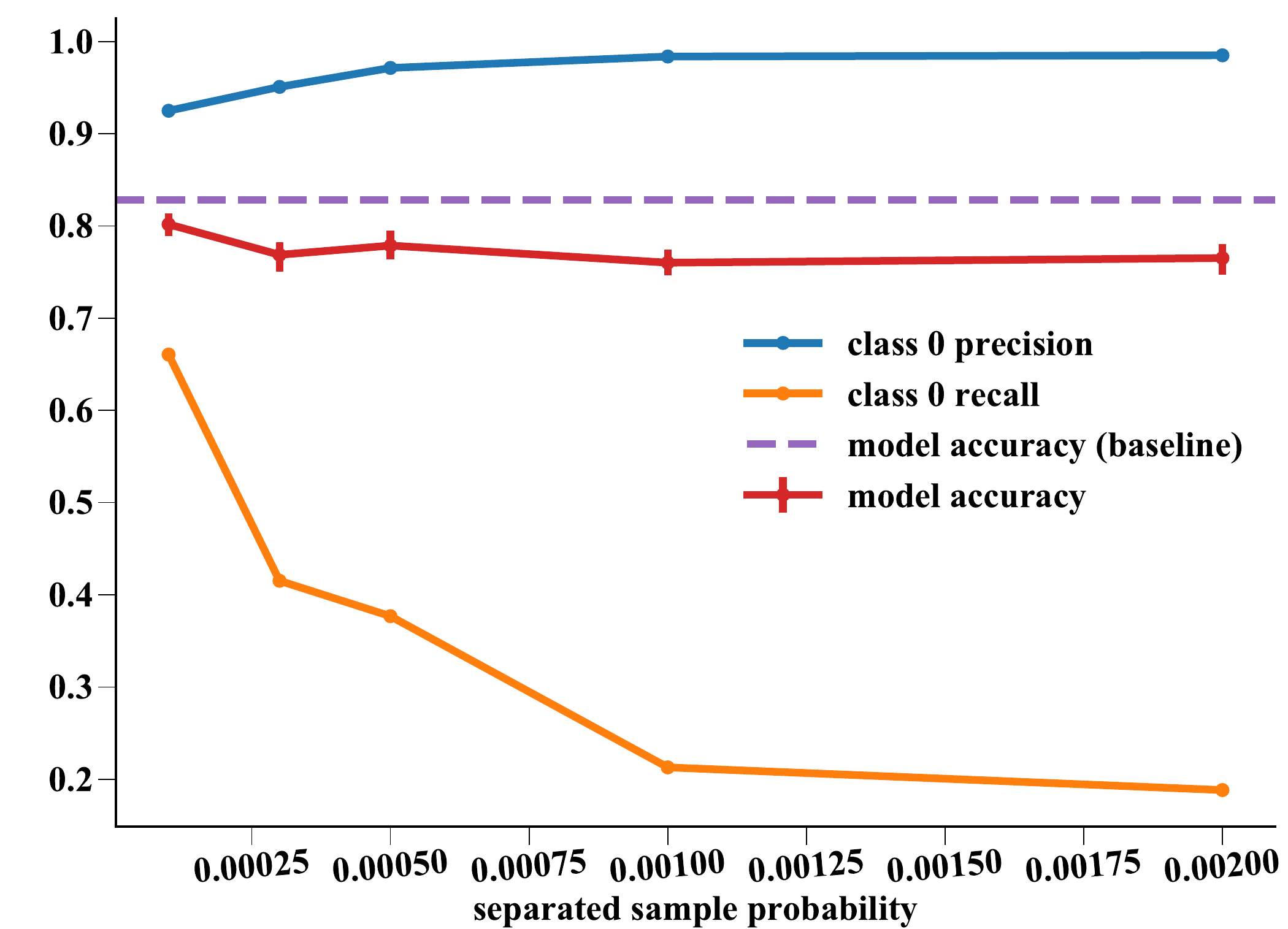}
\end{minipage}

\caption{Attacking recall with different \nsprobs ($\mps$).
The figure on the left shows the results for MNIST;
on the right is CIFAR-10.
All lines are plotted by using average values of five runs.
The model accuracy line also plots the max and min values.
Note that the y-axes are in different scales.
}
\label{fig:c5}

\end{figure*}

\label{ss:eval:recallatk}
\heading{Attacking recall.}
Instead of targeting precision,
\nodeatks can also target class recall.
The rationale is if we can bias the model towards one class by letting
\asgnodes \emph{only see this class}, we should instead bias it away from the
class by letting \asgnodes \emph{only see other classes}.
We test this idea with $\mpn = 10\%$
and let the \asgnodes see all other classes (namely, class 1--9)
with various $\mps$.
Here $\mps=0.01\%$ means that the \asgnodes
will see each class' samples (class 1--9) with probability  0.01\%.
Figure~\ref{fig:c5} shows the results.

From Figure~\ref{fig:c5},
we see that recall of \tcls drops significantly while its precision increases.
This is the reversed effect of previous \nodeatks:
the model predicts \tcls less often and need to be more confident to predict \tcls.
Again, MNIST shows less significant changes compared to CIFAR-10.
We believe this is due to the same reason why previous precision-targeted
attacks do not work as well on MNIST:
MNIST is easier to learn which requires stronger biases and hence
strong attacks (larger $\mpn$ and $\mps$).

\label{ss:eval:bnodeatk}
\heading{\CF{\bnodeatk}.}
As mentioned in section~\ref{ss:bnodeatk},
attackers can conduct attacks blindly
by clustering input tensors and performing a one-shot \nodeatk.
To experiment with this blind attack,
we cluster the input tensors to dropout in each batch
and wait for a cluster of size $\geq 10$.
Then, we use the first 10 samples to conduct a one-shot \nodeatk.
We conduct this attack separately for each epoch (12 epochs in CIFAR-10 training)
to see how accurate clustering is in different epochs.
Table~\ref{table:blind_node_sep_cifar_results} shows the results for CIFAR-10.
Each row in Table~\ref{table:blind_node_sep_cifar_results}
represents a single run that blindly attacks a random class.

\begin{table*}[t]
\caption{
Results of \bnodeatks on CIFAR-10.
The ``attacking epoch'' represents the attack happens in which epoch.
The ``true labels in cluster'' represents the ground truth classes of the cluster seen by the \asgnodes.
The ``target class'' is the most common class in the cluster.
}
\centering
\begin{tabularx}{0.9\textwidth}{@{}c c c c c c  c c@{}}
\toprule

attacking & true labels & target &  model accuracy \%    & \multicolumn{2}{c}{class recall \%} & \multicolumn{2}{c}{class precision \%} \\
epoch     & in cluster  & class  &  (baseline: 82.8) & baseline & attacked     & baseline & attacked  \\

\hline
1  & [5 4 2 6 4 3 4 6 6 6] & 6 & 81.5 & 88.8 & 90.6 & 84.0 & 79.6 \\
2  & [2 8 0 0 0 8 0 0 0 0] & 0 & 81.2 & 84.5 & 89.9 & 84.6 & 76.1 \\
3  & [2 0 0 4 0 0 2 4 0 4] & 0 & 81.5 & 84.5 & 85.8 & 84.6 & 79.8 \\
4  & [0 0 8 8 0 0 0 0 8 0] & 0 & 78.7 & 84.5 & 94.9 & 84.6 & 58.0 \\
5  & [4 5 7 7 7 6 7 7 5 5] & 7 & 81.2 & 85.3 & 87.3 & 88.1 & 85.1 \\
6  & [8 8 8 8 8 8 8 8 8 8] & 8 & 77.7 & 89.3 & 97.7 & 88.3 & 58.2 \\
7  & [5 7 7 4 7 7 7 7 4 7] & 7 & 78.0 & 85.3 & 96.9 & 88.1 & 55.7 \\
8  & [6 6 6 3 6 6 3 6 6 6] & 6 & 75.2 & 88.8 & 97.9 & 84.0 & 45.2 \\
9  & [4 4 4 4 7 4 4 7 7 7] & 4 & 78.6 & 68.1 & 54.7 & 68.5 & 70.9 \\
10 & [9 9 9 9 9 9 9 9 9 9] & 9 & 73.4 & 89.5 & 99.1 & 88.4 & 41.8 \\
11 & [1 1 1 1 8 1 1 1 8 1] & 1 & 77.1 & 91.7 & 98.8 & 92.0 & 59.1 \\
12 & [1 1 1 1 1 1 1 1 1 1] & 1 & 79.2 & 91.7 & 96.1 & 92.0 & 74.2 \\
\bottomrule
\end{tabularx}
\label{table:blind_node_sep_cifar_results}
\end{table*}
 
Since \bnodeatks have too many non-deterministic factors (e.g.,
what samples are contained in the current batch,
a cluster contains what samples,
which cluster has been chosen),
they cannot consistently attack the same class.
Therefore, it is hard to make any qualitative arguments about the attack.
Nonetheless, we can summarize some qualitative trends from
Table~\ref{table:blind_node_sep_cifar_results}.

First, \bnodeatks generally follow the observations that we have seen 
in other \nodeatk variants.
The blind attacks bias the model toward the most dominant class
label in the chosen cluster (the ``target class'' column in Table~\ref{table:blind_node_sep_cifar_results}).
The only exception is in ``epoch 9'', where the most common label in
the cluster is ``class 4'' but the model does not bias toward that class.
Further analysis shows that the model instead biases towards the secondary class,
class 7.

Second, the attacking epoch seemingly matters,
but this is only true in the earliest epochs (epoch $\le 2$ in CIFAR-10).
Using hierarchical Ward clustering~\cite{ward1963hierarchical},
we find that the clusters in our experiments are
good enough to get a dominant class that creates the bias
as early as the second epoch.
While this may differ from dataset to dataset,
we believe that this observation will persist since we only need very few samples from
the same class to perform this attack.
In our current implementation,
we use a simple clustering algorithm, Ward clustering.
We leave exploring other clustering algorithms to future work. 

\section{Related work}
\label{s:relwork}

\heading{Dropout.}
Dropout~\cite{hinton2012improving,srivastava2014dropout}
is a widely used regularization term
designed to overcome overfitting.
A standard dropout implementation drops units uniformly at random.
Dropout also has many variants, including standout~\cite{ba2013adaptive},
fast dropout~\cite{wang2013fast},
variational dropout~\cite{kingma2015variational},
concrete dropout~\cite{gal2017concrete},
and targeted dropout~\cite{gomez2019learning}.
In principle, \atk could apply to these variants as well.
However, for certain variants,
attacking non-determinism while maintaining their semantics can be tricky.
For example, the targeted dropout ranks units and
drop the ``unimportant'' ones stochastically.
It is unclear if \atk is strong enough to create biases
given the number of unimportant units in each batch.
Studying \atk's applicability to major dropout variants
is a topic of future work.

\heading{Training-time attacks.}
Poisoning  attacks against ML at training time have been 
studied extensively in  adversarial machine learning~\cite{gu2017badnets,turner2018clean,liu2017trojaning,
tang2020embarrassingly,biggio2012poisoning,xiao2015feature,
jagielski2018manipulating,shafahi2018poison,jagielski2021subpopulation,
zhao2022towards}.

Poisoning attacks could be classified into: availability attacks that degrade a model's accuracy indiscriminately, targeted attacks that target a few samples at testing time, and backdoor attacks that  misclassify testing samples with a particular backdoor pattern.  Poisoning availability attacks  have been designed for SVMs~\cite{biggio2012poisoning}, linear regression~\cite{xiao2015feature,jagielski2018manipulating}, logistic regression~\cite{mei15teaching}, and neural networks~\cite{lu2022indiscriminate}. Targeted poisoning attacks have been studied in computer vision~\cite{koh2017understanding, shafahi2018poison, suciu2018does,geiping2020witches}, and text models~\cite{schuster2020humpty}.  
Backdoor poisoning attacks~\cite{gu2017badnets,chen2017targeted, turner2018clean,Latent_Backdoors, Wenger2020BackdoorAA, Explanation_Poisoning} alter a small number of training samples to install a backdoor trigger into ML models. Subpopulation poisoning attacks~\cite{jagielski2021subpopulation} target a particular subpopulation in the data distribution and remain stealthy. Trojan attacks~\cite{liu2017trojaning,tang2020embarrassingly} require retraining the model or changing the network architecture to install trojans into the model that can be later activated by an adversary. Most of these attacks require modifications to either the training set, or the model parameters, and might fail under a suite of integrity checks performed on the training data and model parameters. In contrast, \atk only manipulates external randomness used in the training process, and remains undetectable upon inspection of the training set and model parameters. 

We refer the readers to a survey on poisoning attacks in ML~\cite{poisoning_survey} for more related work in this area, as well as the recent NIST report on adversarial ML taxonomy~\cite{advml_report}, which includes Section 4 on poisoning attacks and mitigations.

\heading{Attacking non-determinism in DL training.}
Among the wide range of existing training-time attacks,
there are two that manipulate non-determinism used during model training.
They are the most relevant to \atk.

Asynchronous poisoning attacks~\cite{sanchez2020game}
work in the outsourced training setup, like \atk,
and target asynchronous training (e.g., asynchronous SGD~\cite{zheng2017asynchronous}).
Attackers can reduce model accuracy or bias the model towards a target class
by controlling the scheduling of different asynchronous training threads.
Compared with asynchronous poisoning attacks,
\atk{}s share the same setup
and have similar attack outcomes,
while our attacks target a different non-determinism, dropout operators,
and hence require different attack strategies.

Data ordering attacks~\cite{shumailov2021manipulating}
manipulate the randomness within the SGD optimization process.
By cherry-picking the order in which data is seen by the model,
attackers can slow down the model's learning or let the model learn
some coarse-grained features.
Compared with  data ordering attacks, \atk can control the attack outcome
at finer granularity. For instance, our attacks can tamper with
either precision or recall of a particular target class.

\section{Defense, Limitations, and Discussion}
\label{s:disc}

We discuss here several potential mitigation strategies of \atk, limitations of our work, and avenues for future work. 

\heading{Defending \atk.}
There are three potential avenues for mitigating \atk.
\begin{myitemize2}

\item \emph{Attack prevention:}
One approach to prevent \atk is deploying systems that guarantee execution integrity.
For example, one can run the randomness generator within TEEs
such as SGX, in which the randomly generated outputs are signed by TEE's private
key (unfungible outside the TEE). The training framework will check signatures
before using the randomness. Here, we assume the framework is unmodified, given
our threat model (\S\ref{ss:threat})---only the randomness can be tampered with.
This requires modifying DL frameworks, including splitting and putting the
random generators into TEEs, checking the signatures on the generated random
numbers, and alerting users when detecting violations.

\item \emph{Attack detection:}
We can run verifiable random functions~\cite{micali1999verifiable,dodis2005verifiable}
to generate randomness
that can be cryptographically verified later. A \atk is detected when the
verification fails.

\item \emph{Malfunction detection:} 
The influence of \atk on the model might be detected through diverse testing
methods, statistical analysis, and advanced techniques, such as Meta Neural
Trojan Detection~\cite{xu2021detecting}.
However, these approaches struggle to
pinpoint the root cause of the model malfunction: whether the observed behavior is due to
poor training data, a bug, or an actual attack. If, indeed, the model is under  attack, identifying the specific type of attack is also challenging. In particular, \atk is difficult to detect as the subtle modifications to the dropout layer respect the layer semantics.  

\end{myitemize2}

\heading{Limitation: Conducting \atk in practice.}
In principle, \atk necessitates only the manipulation of the randomness used in dropout layers (\S\ref{ss:threat})
without interfering with other parts of the deep learning framework.
However, our current implementation (\S\ref{s:impl}) requires a level of effort
similar to modifying the DL framework (e.g., conducting supply chain attacks~\cite{supplychainatk,zheng2023careful}).
This is a limitation of our \atk implementation.
Nevertheless, we believe our new threat model is of independent interest,
as the full potential of \atk is beyond our
implementation. For example,  there is a possibility of directly targeting directly Pseudo
Random Number Generators (PRNG) to carry out \atk, a topic that
requires further research.

\heading{\atk on larger models and datasets.}
\atk demonstrates a more profound impact on larger models and datasets.
Our experiments (\S\ref{s:eval}) reveal this trend,
as \atk provides more significant loss and is harder to
recover on larger models.
For example, \nodeatks on VGG16 trained on CIFAR-100 
outperform the smaller models on CIFAR-10
in terms of precision loss,
and in turn, the CIFAR-10 CNN model surpasses  the smaller model on MNIST (Figure~\ref{table:prob_node_sep_results}).
Similarly, larger
models are harder to recover from \maxatks (Figure~\ref{fig:A2_result}) and 
\dropatks (Figure~\ref{fig:B2_result}).
Our hypothesis is that more complex models and dataset offer more ``leeways'' to
\atk because: (1) more diverse inputs and a larger number of classes will
decrease repetitions, increasing the odds that \atk selectively shows
or hides specific signals and information; (2) sophisticated models have larger
capacity, leading to their ability to find a local-minimum that performs well
on the overall model accuracy, and at the same time fails on the target class.

\section{Conclusion}

We introduce \atk, a new family of poisoning attacks
that manipulate non-determinism in the dropout operator.
\atk is stealthy as it does not alter any externally observable states and
can pass today's integrity checks.
\atk is also capable of achieving multiple adversarial objectives. Attackers can slow down (and sometimes even stop) model training,
destroy the prediction accuracy on target classes,
and even sabotage selective metrics (precision or recall) of a targeted class.
By demonstrating \atk,
we want to highlight that \emph{non-determinism is dangerous},
especially in today's outsourced training environment. Mitigating these insidious attacks and verifying the correct operation of randomized  ML training remain important avenues for future work. 
\section*{Acknowledgments}
\vspace{-2ex}
\noindent
We thank our shepherd and the anonymous reviewers of the
IEEE Security and Privacy Symposium 2024 for their feedback that substantially improved this paper. This work was funded by NSF CAREER Awards \#2237295, the ARL Cyber Security CRA under Cooperative Agreement Number W911NF-13-2-0045, and the  DoD Multidisciplinary Research Program of the University Research Initiative (MURI) under contract W911NF-21-1-0322.

\begin{flushleft}
\footnotesize
\setlength{\parskip}{0pt}
\setlength{\itemsep}{0pt}
\bibliographystyle{abbrv}
\bibliography{references}
\end{flushleft}

\newpage

\appendices

\section{Meta-Review}

\subsection{Summary}
This paper presents a new type of attack, DropoutAttack, that targets the
outsourced training setup of deep learning models to drop out specific 
units/classes from the prediction. Using outsourced/third-party servers for 
model training is quite common as it reduces the cost. It works by 
manipulating the dropout operators and breaking the assumption that the 
dropped units are picked randomly. Four variants of DropoutAttack are 
proposed and tested. The evaluation with CIFAR-10 and MNIST datasets showed 
that the proposed attack can reduce the model accuracy by 10-12\% and 
decrease the recall rate to 0\% for the targeted class.

\subsection{Scientific Contributions}
\begin{itemize}
\item Identifies an Impactful Vulnerability
\end{itemize}

\subsection{Reasons for Acceptance}
\begin{enumerate}
\item This paper proposes a new classes of attacks for neural networks 
relying on non-determinism in the dropout process.
\item It is a variant of a supply-chain attack.
\item The proposed vulnerability is impactful and very subtle, and opens
up for new possibilities for attacking neural networks.
\end{enumerate}
 
\end{document}